\documentclass[runningheads]{llncs}

\usepackage[review,year=2024,ID=9617]{eccv}
\usepackage{eccvabbrv}
\usepackage{bbm}
\usepackage{graphicx}
\usepackage{booktabs}
\usepackage{multirow}
\usepackage{tabularray}
\usepackage[dvipsnames]{xcolor}
\usepackage[accsupp]{axessibility}  % Improves PDF readability for those with disabilities.

\usepackage[pagebackref,breaklinks,colorlinks,citecolor=eccvblue]{hyperref}
\usepackage{orcidlink}
\usepackage{multirow} 

\begin{document}
\nolinenumbers
% ---------------------------------------------------------------
% TODO REVIEW: Replace with your title
\title{Beyond Pixels: Semi-Supervised Semantic Segmentation  with a Multi-scale Patch-based Multi-Label Classifier} 

% TODO REVIEW: If the paper title is too long for the running head, you can set
% an abbreviated paper title here. If not, comment out.
\titlerunning{Beyond Pixels}

% TODO FINAL: Replace with your author list. 
% Include the authors' OCRID for the camera-ready version, if at all possible.
\author{Prantik Howlader\inst{1} \and
Srijan Das\inst{2} \and
Hieu Le\inst{3} \and
Dimitris Samaras\inst{1}}

% TODO FINAL: Replace with an abbreviated list of authors.
\authorrunning{P.Howlader et al.}
% First names are abbreviated in the running head.
% If there are more than two authors, 'et al.' is used.

% TODO FINAL: Replace with your institution list.
% \institute{Stonybrook University \and
% \email{\{phowlader; samaras\}@cs.stonybrook.edu}\\
% \url{http://www.springer.com/gp/computer-science/lncs} \and
% ABC Institute, Rupert-Karls-University Heidelberg, Heidelberg, Germany\\
% \email{\{abc,lncs\}@uni-heidelberg.de}}

\institute{Stonybrook University, New York, USA \\
%\email{\{phowlader; samaras\}@cs.stonybrook.edu} 
\and
%\url{http://www.springer.com/gp/computer-science/lncs} \and
University of North Carolina at Charlotte, NC, USA\\
%\email{sdas24@uncc.edu} 
\and
EPFL, Laussane, Switzerland\\
%\email{minh.le@epfl.ch}
}

\maketitle

\begin{abstract}
%Recently, in semi-supervised semantic segmentation (SSS), teacher-student frameworks have shown promising performance with limited labeled data. 
Incorporating pixel contextual information is critical for accurate segmentation. In this paper, we show that an effective way to incorporate contextual information is through a patch-based classifier. This patch classifier is trained to identify classes present within an image region, which facilitates the elimination of distractors and enhances the classification of small object segments. Specifically, we introduce \textbf{Multi-scale Patch-based Multi-label Classifier} (MPMC), a novel plug-in module designed for existing semi-supervised segmentation (SSS) frameworks. MPMC offers patch-level supervision, enabling the discrimination of pixel regions of different classes within a patch. Furthermore, MPMC learns an adaptive pseudo-label weight, using patch-level classification to alleviate the impact of the teacher’s noisy pseudo-label supervision on the student. This lightweight module can be integrated into any SSS framework, significantly enhancing their performance. We demonstrate the efficacy of our proposed MPMC by integrating it into four SSS  methodologies and improving them across two natural image and one medical segmentation dataset, notably improving the segmentation results of the baselines across all the three datasets. Code is available at: \url{https://github.com/cvlab-stonybrook/Beyond-Pixels}.

\end{abstract}

\section{Introduction}
\label{sec:intro}
% Box-classifier generalizes better than pixel-segmenter, especially in low-data regime
% Box-classifier with our box pseudo-label is more robust to noise. 
% We use box-classifier to provide supervision to train a segmentor in a teacher-student framework.
\begin{figure*}
\centering

\includegraphics[width=0.9\linewidth]{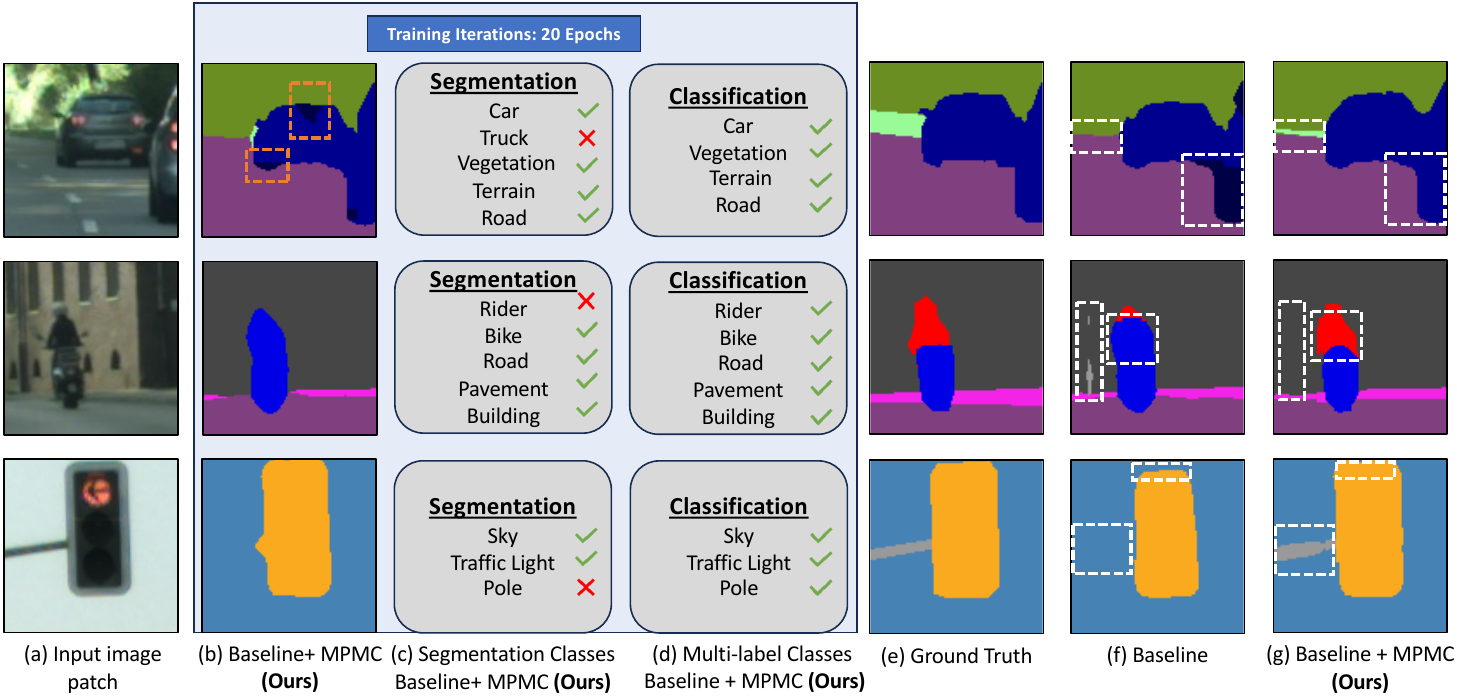}
  \caption{\textbf{Segmentations and Multi-label classifications of image patches (Cityscapes)} After 20 epochs, (b) and (c) are the segmentations and classes in the segmentations respectively, (d) is classes predicted by multi-label classifier in MPMC. (f) and (g) are segmentations by Baseline (AugSeg \cite{zhao2023augmentation})  and Baseline + MPMC at the end of training. White boxes represent regions where MPMC improves the baseline, while orange boxes represent regions where segmentation network predicts classes not present in the patch.
  %(a) input image patches, after initial 20 training epochs (b) are segmentations by baseline + MPMC (Multi-scale Patch-based Multi-label classification), (c) classes predicted by the segmentation network of Baseline + MPMC  and  (d) classes predicted by the MPMC network of Baseline + MPMC, %(e) Ground truths, (f) and (g) are segmentations by Baseline   and Baseline + MPMC at the end of training. Dashed white boxes represent the regions where MPMC improves the baseline. Dashed orange boxes represent the portions where segmentation network predicts classes not present in the patch. The methods are trained on the $\frac{1}{16}$ partition of Cityscapes Dataset. Baseline is AugSeg \cite{zhao2023augmentation}.
}
  \label{fig:teaser}
\end{figure*}
% Minor: receptive field is an important factor when training box-classifier, this also can explain why box-classifier is easier than pixel-classifier. 
Supervised segmentation requires a large amount of pixel level annotations, which is laborious and time-consuming. This paper focuses on semi-supervised segmentation (SSS) \cite{alonso2021semi,hu2021semi}, which alleviates the reliance on labeled annotations by leveraging a large quantity of unlabeled images, accompanied by a limited number of manually labeled images. 

The fundamental task in segmentation is a per-pixel classification. Thus contextual information of a pixel is important to improve its performance \cite{zhou2019context, liu2015parsenet, lin2016efficient}. Context becomes more critical in SSS settings, which uses limited labeled data making the models more prone to overfitting. We note that recent approaches in SSS methods which predominantly leverage consistency regularization with pseudo-labeling techniques~\cite{yang2023revisiting, oliver2018realistic, chen2021semi, french2019semi, hu2021semi, huo2021atso, ouali2020semi, wang2022semi, zou2020pseudoseg} do not exploit contextual information explicitly.

We hypothesize that \textit{patches} are the simplest way to  determine the context around pixels. The key observation is that the contextual information of an image patch are the classes present in it.  For example, in the image patch in the first row of Fig. \ref{fig:teaser} (a) the classes present are car, road, vegetation and terrain. 
Thus, training a network to not only classify each pixel in an image but also predict the classes in the patch surrounding it, is a straight forward way to introduce contextual information in SSS pipeline. Moreover, utilizing multi-scale patch-level supervision can  effectively encapsulate context at varying scales and distinguish pixels across different classes. To this end, we introduce a novel plug-in module, termed as \textbf{M}ulti-scale \textbf{P}atch-based \textbf{M}ultilabel \textbf{C}lassifier (MPMC). This module achieves enhanced discrimination of pixel regions between classes through additional multi-scale patch-level supervision. Notably,  we observe that when training our baseline SSS model together with MPMC in limited labeled data scenario, the patch classifier exhibits greater  resilience to noise than the segmentation network as illustrated in  Fig. \ref{fig:teaser} (b) and (c).  A key observation is when MPMC and baseline SSS model are trained together, the patch classifier provides patch-level labels, identifying the classes present within a region. This facilitates the  elimination of distractors (first row in Fig. \ref{fig:teaser} (b), (c) and (d), while the segmentation network classifies pixels as truck which is not present in the patch, the multi-label classifier does not predict it  ) and enhances the classification of small object segments by the baseline method. %For example in Figure \ref{fig:teaser} first row, after 20 training epochs, MPMC is able to correctly predict the classes in the patch, though the SSS baseline model predicts the class truck even though it is not present in the patch. This leads to the jointly trained segmentation network eliminating distractors when compared to the baseline. %Further in Figure \ref{fig:teaser} third row we observe that after 20 training epochs the patch classifier is able to pick up small objects like pole, not predicted by the segmentation network leading to the jointly trained segmentation network able to segment pole. 

Another important issue plaguing SSS methods is noisy pseudo-labels generated by the teacher for unlabeled images to supervise the student. This can lead to the student model overfitting to these noisy labels, consequently impairing its performance.
Recent approaches have attempted to mitigate the impact of noise in pseudo-labels through various strategies, such as implementing a hard threshold \cite{french2019semi, ke2020guided, ouali2020semi}, applying dynamic re-weighting to enhance the influence of reliable pixels \cite{hu2021semi}, and employing negative data mining by archiving unreliable pixels in a memory bank \cite{wang2022semi}. In contrast to these approaches, we use the  MPMC predictions to learn an adaptive weight that reduces the weight of noisy pseudo-labels in the teacher-student framework. Our approach hinges on the underlying observation that \textit{MPMC exhibits greater uncertainty (lower confidence) for noisy labels in a patch} (see supplementary for details) .

In practice, MPMC is a parameterized module that can be plugged in the initial layers of the visual model, where it leverages the low-level pixel information to classify the classes present in each patch. This patch-level classification strategy forces the visual model to learn and preserve discriminative representations of classes, which are often prone to dilution in the deeper layers of the model.
% By integrating this parameterized module, MPMC, into existing consistency regularization based SSS pipelines~\cite{yang2023revisiting, zhao2023augmentation, }, we improve the effectiveness and generalizability of SSS. 
We demonstrate the efficacy of our proposed parameterized module, MPMC by integrating it into four SSS methodologies—UniMatch \cite{yang2023revisiting}, AugSeg \cite{zhao2023augmentation}, AEL \cite{hu2021semi}, and U2PL \cite{wang2022semi} - notably improving performance across all data partitions in two natural (Cityscapes and Pascal VOC) and one medical (ACDC) dataset. 
\noindent To summarize, our contributions are:
\begin{enumerate}
%\begin{enumerate}
    \item We introduce a novel plug-in module, called MPMC that incorporates an auxiliary multi-scale multi-label classification task, enhancing pixel region discrimination across different classes via additional patch-level supervision.  
    \item MPMC consists of two adaptive learning weights that reduces the influence of noisy pseudo-labels from the teacher model.
    \item We demonstrate MPMC's robustness by integrating in four SSS techniques, achieving consistent improvements across all protocols on Pascal VOC and Cityscapes. Notably, our method also shows improvement in medical image dataset ACDC, which indicates that our method is applicable beyond natural images.
\end{enumerate}

\section{Related Work}

Semi-supervised learning (SSL) has been extensively researched in recent years. Popular SSL approaches include consistency regularization \cite{bachman2014learning,french2019semi,ouali2020semi,sajjadi2016regularization,yang2022st++,durasov2024enabling,durasov2024zigzag}, entropy minimization \cite{chen2021semisupervised,grandvalet2004semi}, and pseudo-labeling \cite{lee2013pseudo,sohn2020fixmatch,Le_CVPRW19,Le_RS22,Le_ICCV2017,Le_ECCV18}. In this work, our emphasis is placed on the domains of pseudo-labeling and consistency regularization.

\noindent
\textbf{Pseudo labeling:}
Pseudo-labeling methods \cite{lee2013pseudo,shi2018transductive} alongside self-training techniques \cite{yarowsky1995unsupervised,mcclosky2006effective} are designed to leverage labeled image data to produce pseudo-labels for unlabeled data. An inherent problem that plagues pseudo-labeling is noise, thus most approaches \cite{zou2020pseudoseg,selvaraju2017grad,sohn2020fixmatch,rizve2021defense} devise strategies to refine the pseudo-labels. PseudoSeg \cite{zou2020pseudoseg}, for instance, aims at enhancing pseudo-label accuracy using attention mechanisms based on grad-cam \cite{selvaraju2017grad}.
In pseudo-label selection, FixMatch \cite{sohn2020fixmatch} adopts a confidence threshold, whereas UPS \cite{rizve2021defense} relies on uncertainty measures.
%FixMatch \cite{sohn2020fixmatch} employs a confidence threshold, and UPS \cite{rizve2021defense} utilizes uncertainty measures for the selection of more reliable pseudo-labels.
ST++ \cite{yang2022st++} focuses on selecting unlabeled images more likely to yield accurate pseudo-pixels.
Emerging approaches like U2PL \cite{wang2022semi} explore the idea of learning  reliable pseudo-labels from their less reliable counterparts.
CFCG \cite{li2023cfcg} introduced a novel approach that combines cross-fusion and contour guidance for pseudo-label improvement.  We differ from these methods by explicitly introducing patch-level contextual information in the SSS methods.

\noindent
\textbf{Consistency Regularization:}
This approach enforces the model to have consistent predictions on different views of the unlabeled images. UniMatch \cite{yang2023revisiting} uses dual perturbations and feature perturbations to force the network to learn more consistent information. ICT \cite{verma2019interpolation} uses Mixup \cite{zhang2017mixup} augmentation for consistency regularization.  Recent methods \cite{french2019semi,yang2022st++,yuan2021simple} use Cutout \cite{devries2017improved}, Cutmix \cite{yun2019cutmix}, Classmix \cite{olsson2021classmix} as strong data augmentation. Consistency Regularization is used with pseudo-labeling techniques in Mixmatch \cite{berthelot2019mixmatch} and TC-SSL \cite{zhou2020time}. In our work, we use a set of strong and weak data augmentations \cite{zhao2023augmentation, alonso2021semi} for generating pseudo-labels.

\noindent \textbf{Multi-label classification:}
This approach classifies an image into more than one label present in the image. Most works in this field explore finding the correlation between the labels and the images \cite{huang2015learning, read2011classifier, yu2014multi}. 
Recent works mainly focus on solving data imbalance issues \cite{guo2021long, wu2020distribution}. We use multi-label classification to help in segmentation. Some works \cite{ge2018multi, reiss2021every} use multi-label classification for segmentation in weakly supervised scenarios, for improved CAM-based pseudo-segmentation. However, we are using multi-label classification in a semi-supervised setting to introduce patch-level contextual information and also use the classifier confidence to reduce the impact of noisy pseudo-labels.

\section{Semi-Supervised Segmentation Pipeline}
Given a labeled set $\mD^{l} = \{(x^{l},y^{l})\}$ and an unlabeled set $\mD^{u} = \{x^{u}\}$, the goal of semi-supervised semantic segmentation is to develop a model that effectively utilizes both labeled and unlabeled data. 
%This section commences with an overview of the foundational framework for semi-supervised segmentation in Section \ref{Preliminary}. Subsequently, an overview of our proposed MPMC, is detailed in Section \ref{Overview}. The section concludes with an elaboration of MPMC's functionality (Section \ref{PMC})
This section commences with an overview of the foundational framework for semi-supervised segmentation in Section \ref{Preliminary}. followed by an overview of our proposed MPMC in Section \ref{Overview}. The section concludes with an elaboration of MPMC's functionality (Section \ref{PMC}).

\begin{figure*}[h]
\centering
 \includegraphics[width=.48\textwidth, height=10\baselineskip]{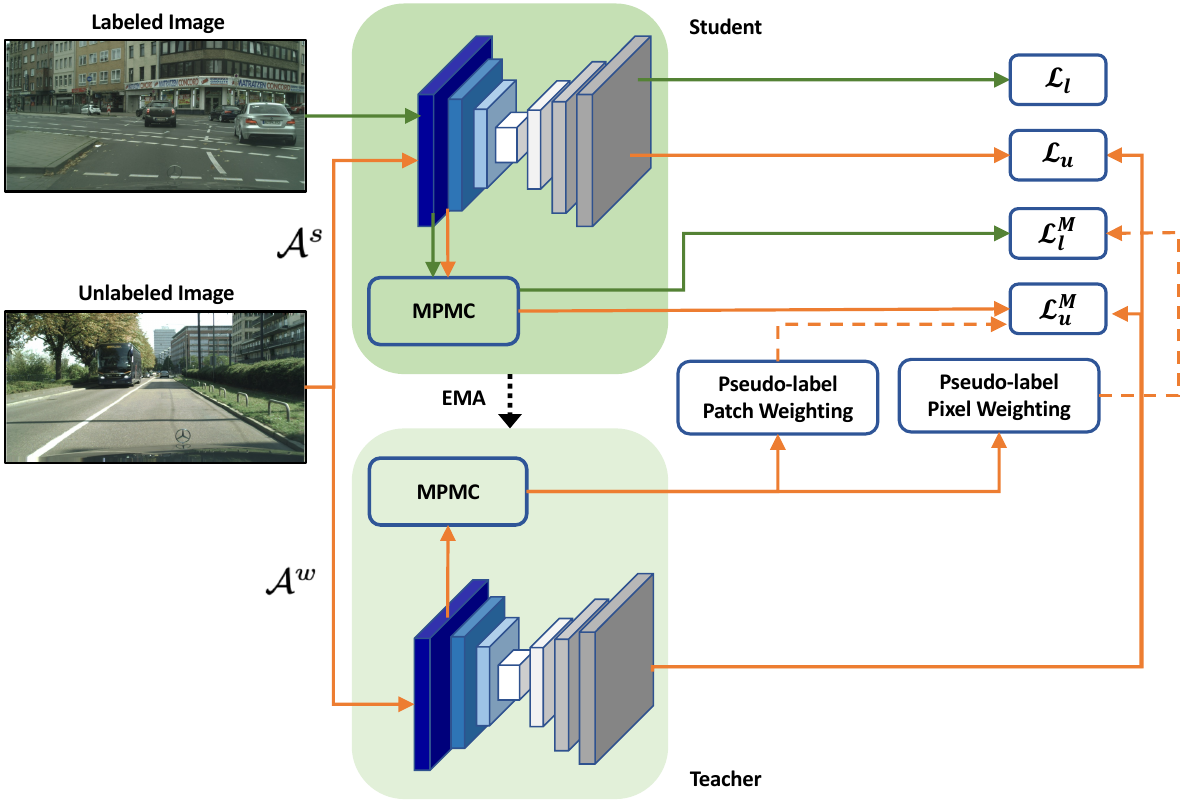}
\rule{1px}{120px}
\hfill
\includegraphics[width=.49\textwidth, height=11\baselineskip]{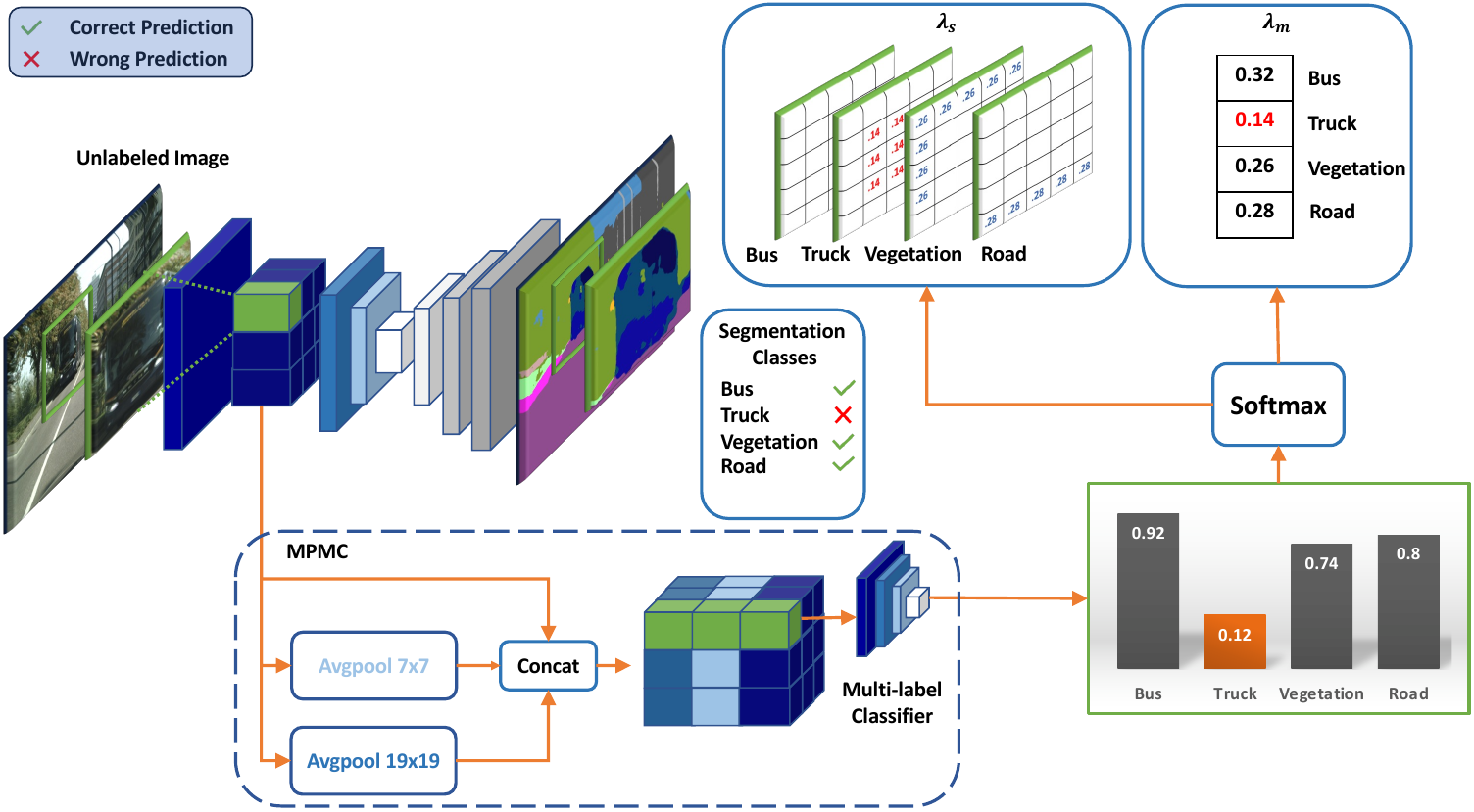}
%\includegraphics[width=.97\textwidth, height=15.1\baselineskip]{pics/multilabel_secondmodel_3.pdf}
% \vspace{0.5cm}
\caption{\textbf{Overall Pipeline of our novel multi-label classification based semantic segmentation:} (a) \textbf{Left}: End-to-end Teacher-Student Pipeline with our novel method Multi-scale Patch-based Multi-label Classifier (MPMC). (b) \textbf{Right}: For unlabeled images, The teacher MPMC extracts features from a layer in the segmentation model's encoder to classify the feature's receptive field patch. The confidence of MPMC for a class in a patch is used to calculate two adaptive weights $\lambda_s$  and $\lambda_m$ which is used to reduce the influence of noisy predictions in that patch from the teacher to train both the student  segmentation network and MPMC.
}
  \label{fig:model}
\end{figure*} 
\subsection{Preliminary}\label{Preliminary}
Teacher-Student frameworks leverage weak-to-strong consistency regularization to leverage unlabeled data. As depicted in Fig. \ref{fig:model}(a), each unlabeled image $x^{u}$ is simultaneously perturbed by two operators, i.e., weak perturbation $\mA^{w}$ (cropping, random flip, etc), and strong perturbation $\mA^{s}$ (color jitter, gaussian blur, contrast,  etc). The framework's objective function is a convex combination of supervised loss $\mL_{l}$ for the labeled data and unsupervised loss $\mL_{u}$ for the unlabeled data as:
\begin{equation}\label{eq1}
    \mL = \mL_{l} + \alpha\mL_{u}  \hspace{0.2in} \textrm{where,} \hspace{0.1in} \mL_{u} = \frac{1}{|\mD^{u}|}\sum \mathbbm{1}(p^{w}\ge th)\mE(p^{w}, p^{s})
\end{equation}
where, $\mathcal{L}_{l}$ is the standard cross-entropy loss, comparing the model's predictions with ground truth. While $\mathcal{L}_{u}$ aims to regularize the predictions for an unlabeled sample so that they remain consistent under both weak and strong augmentations ($p^{w}$ and $p^{s}$). 
Here, %$\mN_{u}$ is the number of unlabeled images, %$p^{w}$ and $p^{s}$ denote the logits of the unlabeled image under weak and strong augmentations respectively. 
$th$ is a predefined confidence threshold to filter noisy pseudo labels and $\mE$ minimizes the entropy between two probability distributions.

\subsection{Overview}\label{Overview}
Fig. \ref{fig:model}(a) illustrates the comprehensive training process of our proposed MPMC, %, outlining a structured approach divided into two key components: 1) A framework consisting of a teacher model for generating pseudo-labels and a student model for online learning; 2) 
which is tailored to enhance the discrimination of patches among different classes and to mitigate the impact of noisy predictions transmitted from the teacher to the student. MPMC can be integrated into both teacher and student models within the teacher-student framework. It is composed of a classifier that performs multi-label classifications of the multi-scale patch representations of the input image. The application of MPMC spans two distinct scenarios: 1) For labeled images, MPMC assigns classes to each patch representation based on the classes present within a patch. The multi-scale nature of these patch representations and their classification enhances the segmentor's ability to discriminate classes by capturing enhanced contextual information related to variations in shape and size. %This auxiliary task of patch classification augments the network's proficiency in learning and preserving class representations, a critical function particularly pertinent as the fidelity of these representations tends to diminish in the network's deeper layers.
2) In the case of unlabeled images, besides patch classification, MPMC computes a pseudo-label weight map to alleviate the impact of the teacher's noisy pseudo-label supervision on the student. 
This involves assigning weights to the pseudo-labels within each patch in consistency learning framework, which are determined by the confidence  of the MPMC's predictions for each class in the patch, to enhance the reliability and accuracy of the pseudo-labeling process.
%Recognizing that a mere confidence-based threshold on the segmentation network's predictions is insufficient for filtering out  misclassified pseudo-labels as the network in the initial training epochs suffers from miscalibration \cite{rizve2021defense} which leads to  an inaccurate student segmentor. Therefore, we propose a novel approach which augments the conventional confidence based pseudo-labeling approach by assigning weights to the confidence-thresholded pseudo-labels of each class in a patch, based on the confidence levels of the multi-label classifier for the corresponding class in that patch. %This strategy is motivated by the fact that a multi-label classifier will exhibit more uncertainty (lower confidence) for noisier predictions.

\subsection{Multi-scale Patch-based Multi-label Classifier (MPMC)}\label{PMC}
%While current semi-supervised methods\cite{alonso2021semi, hu2021semi, yang2023revisiting} show proficiency across various datasets, they often overlook the critical challenge of distinguishing classes that exhibit patch-level similarity, particularly when some of these classes are less represented or 'tail' classes. The tendency of models to favor classes with more labeled instances often results in the misclassification of tail classes in favor of visually similar, more prevalent classes. Hu \emph{et al}.\cite{hu2021semi} suggested enhancing tail class performance by augmenting their labeled instances through a ``copy-paste" method. However, this approach falls short in diversifying the distribution of these tail classes. 

Our proposed MPMC module is specifically designed to address the shortcomings of SSS pipeline. %that arise due to visual similarity of pixels across classes and data imbalance in the training distribution.
%It enhances the discrimination between pixel regions of different classes and mitigates the effect of noisy teacher predictions that arise due to data imbalance within a consistency regularization framework. 
MPMC can be plugged in at any layer within the teacher-student network, both within the teacher model, denoted as $g$, and within the student model, denoted as $\hat{g}$, they share the same parameters.
Below, we detail the training of MPMC on labeled images and then on unlabeled images.
%Concretely, we perform multi-label classification for the patch representations of labeled images, thereby compelling the segmentation model to learn more distinctive features for each class (Section \ref{PMC on Labeled Images}). Further, we introduce a pseudo-label weight map for unlabeled images to reduce the effect of the teacher's noisy pseudo-labels on the student. The weight of a thresholded pseudo-label of a class in a patch by the teacher is based on the confidence of the multi-label classifier for the class(Section \ref{PMC on Unlabeled images}).
% \begin{figure*}
% \centering
% %\includegraphics[width=0.9\linewidth,height=0.23\linewidth]{pics/teaser.pdf}
% \vspace{-1cm}
% \includegraphics[width=1\linewidth]{pics/label_accuracy_area.pdf}
% \vspace{-3cm}
%   \caption{\textbf{Pixel accuracy for different resolutions of PMC classification} pixel accuracy with respect to area of instances for (a) bus and truck, (b)  traffic light and rider. The methods are trained on $\frac{1}{16}$ partition of Cityscapes Dataset.
% }
%   \label{fig:labelaccuracy}
% \end{figure*}
\subsubsection{MPMC on Labeled Images.}\label{PMC on Labeled Images}
%HIEU.... 
%To incorporate a patch-supervision into the segmentation framework, we train a classifier that takes as input a single pixel in the feature map of the $\tau^{th}$ layer of the encoder. This single pixel-representation encodes the information from a patch in the original image. Further, we perform everage-pooling on this features map with different sizes before exactracting the features, thus obtainting multi-scale ...
%%%%%%%%%%%%%%%%%%%%%%%%%%%%%%%%%%%%%%%%%%%%%%%%%%%%%%%%%%%%%%%%%%%%%%%%%%%%%
As illustrated in Fig. \ref{fig:model}, MPMC incorporates patch-supervision into the segmentation framework, by training a multi-label classifier that takes as input a single pixel in the feature map of the $\tau^{th}$ layer of the encoder. This single pixel-representation encodes the information from a patch (receptive field)  in the original image. Further, to encapsulate context at varying scales we perform average pooling on this features map with kernels of different sizes before performing multi-label classification.
% As illustrated in Figure \ref{fig:model}, we train MPMC on the receptive field of the $\tau^{th}$ layer of the encoder, which may encapsulate more than one class. The receptive field represents the patches in the image, thus providing a robust design choice to implement MPMC. %Further, a challenge in SSS task is existence of large intra-class variation, i.e., regions belonging to the same class may exhibit a very different appearances even in the same picture. For example, classes like ``Car'' and ``Bus'' are  represented by instances which show high variability in shapes and sizes. Thus we further propose a multi-scale PMC to better capture different classes i.e., we use average pooling to capture instances at different resolutions to capture more ``context'' in classifying the classes in a patch.

Let us define a patch of image and the corresponding segmentation in a receptive field $r$ as $\{x_{r}^{l},y_{r}^{l}\} \in \mD^l$. 
Note that $y_{r}^{l}$ is a multi-label target where the classes present in the receptive field $r$ are positive classes, and the classes not present are negative classes. 

The input to the student MPMC $\hat{g}$ are the features extracted from the $\tau^{th}$ layer of the encoder in  student segmentation network $\hat{f}$, denoted by $z_r = \hat{f}_{\tau}(\mA_w(x^l_r))$. MPMC captures the contextual information within a patch at different scales, $S_i$, $i \in \{0,1\}$, where $s_0$ and $s_1$ are average pool kernel sizes. For all our experiments $s_0$ and $s_1$ are set to $7 \times 7$ and $19 \times 19$. These pooling operations are performed with appropriate padding to maintain uniformity in the shape of the output features. MPMC $\hat{g}$ is composed of the concatenation of these multi-scale pooled features, followed by a convolutional network $\mM$ and a final linear layer. Thus, the output of the MPMC $\hat{g}$ for the input patch $x^l_r$ is given by:
\begin{equation}\label{eq3}
    q^l = \sigma(W^T\mM(concat(z_r, {z^{S_{0}}_r},{z^{S_{1}}_r})) + b)
\end{equation}
% Then, the classifier $g$ is trained with multi-label supervision $\mL^{M}_{l}$ as:
% \begin{equation}\label{eq3}
%     \mL^{M}_{l} = \frac{1}{|D^{l}|} \sum_{(x^{l},y^{l}) \in D^{l}} \sum^{R}_{r=1}l^M_r
% \end{equation}
where  $W$ and $b$ are linear layer weights and bias, $\sigma$ is the sigmoid activation and $q^l = [q_1^l, ..., q_C^l]^T \in \mathbbm{R}^{C}$ are the predicted probabilities for $C$ classes.

\iffalse
%where $F_{r} \in \mathbbm{R}^{C \times d}$ are the feature vectors  for $C$ categories. To perform multi-label classification, we treat the prediction of each label as a binary classification task. We project feature of each label $F_{r,c} \in \mathbbm{R}^{d}$ to a logit value using a linear projection layer followed by a sigmoid function:
\begin{equation}\label{eq4}
    p_{r,c} = Sigmoid(W^T_c F_{r,c} + b_c)
\end{equation}
Where $W_c \in \mathbbm{R}^{d}$, $W = [W_1, ..., W_C]^T \in \mathbbm{R}^{C \times d}$, and $b_k \in \mathbbm{R}$, $b = [b_1,...,b_C]^T \in \mathbbm{R}^{C} $ are parameters in the linear layer, and $p = [p_1, ..., p_C]^T \in \mathbbm{R}^{C}$ are the predicted probabilities for each class.
\fi 

While binary cross-entropy loss is applicable to train MPMC, it indiscriminately weighs all positive and negative classes equally. Considering the receptive field of a feature from the $\tau^{th}$ layer of the encoder in the segmentation network $\hat{f}$ covers only a small patch relative to the input image's size, the prevalence of negative classes within a patch significantly exceeds that of positive classes. This imbalance influences the binary cross-entropy loss to favor negative classes~\cite{wu2020distribution}. Inspired by long-tailed multi-label classification~\cite{wu2020distribution, 10.1609/aaai.v37i2.25244} which mitigate such imbalances by diminishing the weight of negative classes, we adapt an asymmetric loss \cite{ridnik2021asymmetric}, a focal loss variant that assigns distinct $\gamma$ values to positive and negative classes. %We adopt asymmetric loss \cite{ridnik2021asymmetric}, which is a variant of focal loss with different $\gamma$ values for positive and negative classes. 
For an image patch $x^{l}_r$ and the predicted class probabilities $q^l$ using our framework, the asymmetric focal loss is expressed as:
\begin{equation}\label{eq5}
    l^M_r = \frac{1}{C}\sum_{c=1}^{C}\begin{cases}
    (1-q_c^l)^{\gamma^{+}}log(q_c^l), & y^l_{r,c}=1.\\
    (q_c^l)^{\gamma^{-}}log(1-q_c^l), & y^l_{r,c}=0.
  \end{cases}
\end{equation}
In our experiments, we set $\gamma^{+} = 0$ and $\gamma^{-} = 1$.
Subsequently, MPMC $\hat{g}$ undergoes training with multi-label supervision $\mL^{M}_{l}$, formulated as:
\begin{equation}\label{eq6}
    \mL^{M}_{l} = \frac{1}{|D^{l}|} \sum_{(x^{l},y^{l}) \in D^{l}} \sum^{R}_{r=1}l^M_r
\end{equation}
where R is the number of receptive fields at $\tau^{th}$ layer of the segmentor. This auxiliary task of patch classification augments the network's proficiency in learning and preserving class features, a critical function particularly pertinent as the fidelity of these features tends to diminish in the network's deeper layers.

\subsubsection{MPMC on Unlabeled images.}\label{PMC on Unlabeled images}

In semi-supervised teacher-student framework, for unlabeled images, the prediction of the teacher network $f$ for the weak augmented view $\mA^w(x^u)$ is used to supervise the student network $\hat{f}$ for the strong augmented view $\mA^s(x^u)$. 
In the early iterations of semi-supervised learning, the teacher's predictions are inherently noisy. Common practices \cite{sohn2020fixmatch, zuo2021self, liu2022perturbed, fan2022ucc, hu2021semi}, involve employing pixel-wise confidence score thresholds ($th$) to filter out unreliable pseudo labels. However, a high confidence threshold does not entirely eliminate noise due to the teacher's initial miscalibration~\cite{rizve2021defense}, allowing some noisy predictions to have high confidence. To refine this process, our approach does not solely depend on confidence-thresholded teacher segmentation predictions for reliable pseudo-label extraction. Instead, we leverage the teacher MPMC $g$'s confidence for a patch to adjust the weights of the deemed reliable pseudo-labels within that patch. The underlying principle of our methodology is that \textit{MPMC exhibits greater uncertainty (lower confidence) for noisy labels}.
%, providing a more nuanced mechanism to enhance label reliability for supervising both the student segmentation network $\hat{f}$ and PMC $\hat{g}$.

We employ the teacher MPMC $g$'s predictions to define two adaptive learning weights $\lambda_{m} \in \mathbbm{R}^{R \times C}$ and $\lambda_{s} \in \mathbbm{R}^{W \times H}$, where $R$ denotes the number of patches, sized according to the receptive field within the image, $W$ and $H$ represents the width and height of the input image, respectively. The learning weight $\lambda_{m}$ is utilized to assign 
class weights to each patch for training the student MPMC, whereas $\lambda_{s}$ is applied to diminish the impact of noisy predictions during training the student segmentation network.

For an unlabeled image $x^u$, consider a patch at the receptive field $r$, denoted as $x^u_r$. The teacher and student segmentation networks yield predictions $p^w = f(\mA^w(x^u_r))$ and $p^s = \hat{f}(\mA^s(x^u_r))$, respectively. Simultaneously, the teacher and student MPMC predict $ q^u_r = g(f_{\tau}(\mA^w(x^u_r)))$ and $\hat{q}^u_r = \hat{g}(\hat{f}{\tau}(\mA^s(x^u_r)))$ using features from the $\tau^{th}$ layer of the encoder. Let us define the confidence of teacher MPMC $g$ across all classes in patch $r$ as $\gamma = \textrm{softmax}(q_r^u)$, with $\gamma \in \mathbbm{R}^C$. 
The weight of the patch $r$, $\lambda_{m,r}$ is determined by $\gamma$. For each pixel in patch $r$ associated with class $c$, $\lambda_{s}$ is assigned the weight $\gamma_{c}$. This approach assigns weights to pseudo-labels and pixels based on the teacher MPMC's confidence, enhancing the accuracy of the learning process.

Thus, the control parameters $\lambda_s$ and $\lambda_m$ are integrated to the unsupervised losses to train the student segmentation network and MPMC. The unsupervised segmentation loss $L_u$ mentioned in equation~\ref{eq1} to train the student segmentation network $\hat{f}$ is  re-formulated as:
% \begin{equation}\label{eq8}
% \mL_{u} = \frac{1}{|\mD_{u}|} \sum_{i=1}^{|\mD_{u}|}\frac{1}{WH} \sum_{j=1}^{WH} \mathbbm{1}(max(f( x^{w,u}_{ij} )) \ge th)l_{ij}^u
% \end{equation}
\begin{equation}\label{eq8}
\mL_{u} = \frac{1}{|\mD^{u}|} \sum_{x^{u} \in \mD^{u} } \mathbbm{1}(p^w \ge th)\lambda_{s}\mE(p^w, p^s)
\end{equation}
% \begin{equation}\label{eq9}
%  l^{ij}_u = \lambda_{s}^{ij}\text{CE}(\hat{y}_{ij}^{u},\hat{f}( x^{s,u}_{ij} ))
% \end{equation}
where $\mE$ is the standard cross entropy loss and $th$ is the predefined segmentation confidence threshold. 
Finally, the unsupervised multi-label loss $\mL^M_u$ to train the student MPMC $\hat{g}$ is given by:
\begin{equation}\label{eq10}
    \mL^{M}_{u} = \frac{1}{|\mD^{u}|}\sum_{x^{u} \in D^{u}}\sum_{r=1}^{R}\frac{1}{C}\sum_{i=1}^{C}\lambda_{m,r}^{i}\mE^+(q^u_r, \hat{q}^u_r)  
\end{equation}
\iffalse
\begin{equation}\label{eq11}
l^{M}_{r,i} = - \lambda_{m,r}^{i}[(y_{r}^{u,m})_i \log(\hat{Q}_i) + (1 - (y_{r}^{u,m})_i ) \log(1 - \hat{Q}_i)]
\end{equation}
\fi
where $\mE^+$ is the binary cross entropy loss.
%$y_{r}^{u,m}$ is the multi-label target for the patch $r$ derived from the confidence thresholded prediction of the teacher $f(x^{w,u}_r)$ for that patch.
Therefore, the overall loss function is defined as:\\
\begin{equation}\label{eq12}
    \mL = \mL_{l} + \mL_{l}^M + \alpha\mL_{u} +\beta\mL_{u}^M
\end{equation}
where $\alpha$ and $\beta$ controls the contribution of $\mL_{u}$ and $\mL_{u}^M$. In all our experiments $\alpha$ and $\beta$ are set to $0.1$ and $0.25$ respectively.
\section{Experiments}
\subsection{Setup}
\noindent
\textbf{Datasets:} \textbf{PASCAL VOC 2012} \cite{everingham2010pascal} is a standard semantic segmentation dataset with 20 semantic classes and a background class. The training, validation, and testing sets consist of $1464$, $1449$, and $1556$ images (\textit{Classic}), respectively. Following \cite{zhao2023augmentation, chen2021semi}, we also include the additional augmented dataset \cite{hariharan2011semantic} (\textit{Blender}), which includes a collection of $10582$ training images. We adopt the same partition protocols in \cite{zhao2023augmentation, chen2021semi} to evaluate our method in both \textit{Classic} and \textit{Blender} sets. \textbf{Cityscapes}\cite{cordts2016cityscapes} dataset is designed for urban scene understanding, consisting of 30 classes,  of which only 19 classes are used for scene parsing evaluation. It has $2975$, $500$, and $1525$ images for training, validation, and testing, respectively. \textbf{ACDC}\cite{bernard2018deep} dataset contains 200 annotated short-axis cardiac cine-MR images from 100 patients. The segmentation masks consists of classes:  left ventricle (LV), myocardium (Myo), and right ventricle (RV). 140 images from 70 patients, 60 images from 30 patients are randomly selected for training and validation respectively.

\textbf{Implementation Details:}
For fair comparison, we adopt DeepLabv3+ \cite{chen2018encoder} as the decoder in all of our experimental setups, and compare with both ResNet-101 and ResNet-50 \cite{he2016deep} as the backbone architecture. In all our experiments with Cityscapes, Pascal, and ACDC datasets, the training resolution is 801, 513, and 224 respectively. Note, for maintaining fair comparison, our method ``UniMatch \cite{yang2023revisiting} + MPMC'' uses the training resolution of 321 which is used by UniMatch \cite{yang2023revisiting} when training on the Classic set of Pascal VOC. Further when training a MPMC integrated baseline, we use the same weak and strong augmentations as used by the corresponding baselines.
The convolutional network  in MPMC is comprised of three convolution blocks. Each convolution block consists of a stack of two $3 \times 3$ convolutions followed by a $1 \times 1$ convolution. For all our experiments MPMC is integrated to the first layer of the ResNet Encoder.

\subsection{Comparison with State-of-the-Art Methods}
We conducted our experiments on PASCAL VOC 2012 and Cityscapes which are two popular natural image benchmarks, and a medical dataset ACDC. MPMC is integrated in four semi supervised methods UniMatch \cite{yang2023revisiting}, AEL \cite{hu2021semi}, U2PL \cite{wang2022semi} and AugSeg \cite{zhao2023augmentation}. Note, UniMatch \cite{yang2023revisiting} is a consistency regularization based method. We observe that MPMC consistently improves over their baseline performance across all partition protocols on all datasets. \\%Which strongly proves the effectiveness of our method.\\ 
\textbf{Results on PASCAL VOC 2012 Dataset.} Table \ref{tab:voc} shows the comparison with other SOTA methods for both the \textit{Classic} and \textit{Blender} sets. MPMC integrated to the four  baseline methods improves them consistently across all data partitions for both \textit{Classic} and \textit{Blender sets}. We find that our method brings the most improvement for the least labeled data partition ($\frac{1}{16}$), improving AugSeg, AEL, U2PL and UniMatch by $\textbf{1.8}\%$, $\textbf{2.5}\%$, $\textbf{2.0}\%$ and $\textbf{2.1}\%$ respectively in \textit{Classic} with a ResNet-101 encoder. \\
\textbf{Results on Cityscapes Dataset.} Table \ref{tab:ct} shows the comparison with other SOTA methods. MPMC integrated to the four  baseline methods improves them consistently across all data partitions. We observe that similar to results in PASCAL VOC 2012 dataset, our method brings the most improvement for the least labeled data partition ($\frac{1}{16}$), improving AugSeg, AEL, U2PL and UniMatch by $\textbf{1.7}\%$, $\textbf{2.1}\%$, $\textbf{2.3}\%$ and $\textbf{1.9}\%$ repectively in \textit{Classic} for ResNet-101 encoder.\\
  %Also, its ability to integrate across different SSS methods demonstrates the robustness of our proposed PMC. \\
\textbf{Results on ACDC Dataset.} Table \ref{tab:medical} shows the comparison with other SOTA methods. We observe that when MPMC is integrated to the four  baseline  methods, it significantly improves the performance for the ``3 cases'' data partition which has the least labeled images ($<10\%$ of labeled images). Note we use ResNet-101 for all experiments in this dataset.\\
The consistent improvement in SSS performance in both natural and medical datasets and its ability to integrate in different SSS methods substantiates the robustness of MPMC. Moreover, the notable improvements achieved in scenarios with limited data shows MPMC's vital efficacy within SSS pipelines.

\begin{table*}[]
\begin{subtable}{\linewidth}
\centering
\resizebox{0.98\linewidth}{!}{%
\begin{tabular}{cr|ccccc|ccc}
\hline
\multicolumn{2}{c|}{\multirow{2}{*}{Method}} & \multicolumn{5}{c|}{\textit{Classic}} & \multicolumn{3}{c}{\textit{Blender}} \\ \cline{3-10} \multicolumn{2}{c|}{} & 1/16 & 1/8 & 1/4 & 1/2 & Full & 1/16 & 1/8 & 1/4 \\ \hline\hline
\multicolumn{10}{l}{\textbf{ResNet-50}} \\\hline
\textit{Supervised} &&  44.0  & 52.3  & 61.7  & 66.7  & 72.9  &- & - & - \\
PC$^2$Seg \cite{zhong2021pixel} &{\color[HTML]{656565} {\small[}CVPR'21{]}}&  56.9  & 64.6  & 67.6  & 70.9  & 72.3  & - & - & - \\
PseudoSeg \cite{zou2020pseudoseg} &{\color[HTML]{656565} {\small[}ICLR'21{]}}&  56.9 & 64.6 & 67.6 & 70.9 & 72.2 &- &- &- \\
ST++ \cite{yang2022st++}      &{\color[HTML]{656565} {\small[}CVPR'22{]}}&  -     & -     & -     & -     & -     & 72.6      & 74.4      & 75.4      \\\hline

AugSeg \cite{zhao2023augmentation} & {\color[HTML]{656565} {\small[}CVPR'21{]}} &  64.2 & 72.1 & 76.1 & 77.4 & 78.8 & 77.2      & 78.2      & 78.2 \\
AugSeg \cite{zhao2023augmentation} + MPMC &&  \textbf{65.9} & \textbf{73.7} & \textbf{77.6} & \textbf{78.5} & \textbf{79.7} & \textbf{79.0} & \textbf{79.8} & \textbf{79.7} \\
({\color[HTML]{963400}$\Delta$}) &&  ({\color[HTML]{963400}+1.7}) & ({\color[HTML]{963400}+1.6}) & ({\color[HTML]{963400}+1.5}) & ({\color[HTML]{963400}+1.1}) & ({\color[HTML]{963400}+0.9}) & ({\color[HTML]{963400}+1.8}) & ({\color[HTML]{963400}+1.6}) & ({\color[HTML]{963400}+1.5}) \\
\hline

AEL \cite{hu2021semi} \hfill & {\color[HTML]{656565} {\small[}NeurIPS'21{]}} &  62.9 & 64.1 & 70.3 & 72.7 & 74.0 & 74.1 & 76.1 & 77.9 \\
AEL \cite{hu2021semi} + MPMC && \textbf{65.0} & \textbf{66.1} & \textbf{72.1} & \textbf{74.2} & \textbf{75.3} & \textbf{76.1} & \textbf{78.0} & \textbf{79.6} \\
({\color[HTML]{963400}$\Delta$}) &&  ({\color[HTML]{963400}+2.1}) & ({\color[HTML]{963400}+2.0}) & ({\color[HTML]{963400}+1.8}) & ({\color[HTML]{963400}+1.5}) & ({\color[HTML]{963400}+1.3}) & ({\color[HTML]{963400}+2.0}) & ({\color[HTML]{963400}+1.9}) & ({\color[HTML]{963400}+1.7}) \\
\hline

U2PL \cite{wang2022semi} \hfill & {\color[HTML]{656565} {\small[}CVPR'22{]}} &  63.3 & 65.5 & 71.6 & 73.8 & 75.1 & 74.7 & 77.4 & 77.5 \\
U2PL \cite{wang2022semi} + MPMC &&  \textbf{65.2} & \textbf{67.4} & \textbf{73.3} & \textbf{75.2} & \textbf{76.2} & \textbf{76.5} & \textbf{79.1} & \textbf{79.1} \\
({\color[HTML]{963400}$\Delta$}) &&  ({\color[HTML]{963400}+1.9}) & ({\color[HTML]{963400}+1.9}) & ({\color[HTML]{963400}+1.7}) & ({\color[HTML]{963400}+1.4}) & ({\color[HTML]{963400}+1.1}) & ({\color[HTML]{963400}+1.8}) & ({\color[HTML]{963400}+1.7}) & ({\color[HTML]{963400}+1.6}) \\
\hline

UniMatch \cite{yang2023revisiting}  \hfill & {\color[HTML]{656565} {\small[}CVPR'23{]}} &  71.9 & 72.5 & 76.0 & 77.4 & 78.7 & 78.1 & 79.0 & 79.1 \\
UniMatch \cite{yang2023revisiting} + MPMC &&  \textbf{73.5} & \textbf{74.0} & \textbf{77.4} & \textbf{78.5} & \textbf{79.4} & \textbf{79.6} & \textbf{80.4} & \textbf{80.4} \\
({\color[HTML]{963400}$\Delta$}) &&  ({\color[HTML]{963400}+1.6}) & ({\color[HTML]{963400}+1.5}) & ({\color[HTML]{963400}+1.4}) & ({\color[HTML]{963400}+1.1}) & ({\color[HTML]{963400}+0.7}) & ({\color[HTML]{963400}+1.5}) & ({\color[HTML]{963400}+1.4}) & ({\color[HTML]{963400}+1.3}) \\
\hline\hline
\multicolumn{10}{l}{\textbf{ResNet-101}} \\\hline
\textit{Supervised} &&  45.1 & 55.3 & 64.8 & 69.7 & 73.5 & 70.6 & 75.0 & 76.5 \\
CPS  \cite{chen2021semi} & {\color[HTML]{656565} {\small[}CVPR'21{]}}&  64.1 & 67.4 & 71.7 & 75.9 & - & 72.2 & 75.8 & 77.6 \\
PS-MT \cite{liu2022perturbed} &{\color[HTML]{656565} {\small[}CVPR'22{]}}&  65.8 & 69.6 & 76.6 & 78.4 & 80.0 & 75.5 & 78.2 & 78.7 \\
PCR \cite{xu2022semi} &{\color[HTML]{656565} {\small[}NeurIPS'22{]}}&  70.1 & 74.7 & 77.2 & 78.5 & 80.7 & 78.6 & 80.7 & 80.8 \\
DAW  \cite{sun2023daw} & {\color[HTML]{656565} {\small[}NeurIPS'23{]}} &  74.8 & 77.4 & 79.5 & 80.6 & 81.5 & 78.5 & 78.9 & 79.6 \\
CFCG  \cite{li2023cfcg} &{\color[HTML]{656565} {\small[}ICCV'23{]}}&  - & - & - & - & - & 77.4 & 79.4 & 80.4 \\\hline

AugSeg \cite{zhao2023augmentation} & {\color[HTML]{656565} {\small[}CVPR'21{]}} &  71.1 & 75.5 & 78.8 & 80.3 & 81.4 & 79.3 & 81.5 & 80.5 \\
AugSeg \cite{zhao2023augmentation} + MPMC &&  \textbf{72.9} & \textbf{77.8} & \textbf{79.9} & \textbf{81.3} & \textbf{82.1} & \textbf{80.8} & \textbf{82.5} & \textbf{81.4} \\ 
({\color[HTML]{963400}$\Delta$}) &&  ({\color[HTML]{963400}+1.8}) & ({\color[HTML]{963400}+2.3}) & ({\color[HTML]{963400}+1.1}) & ({\color[HTML]{963400}+1.0}) & ({\color[HTML]{963400}+0.7}) & ({\color[HTML]{963400}+1.5}) & ({\color[HTML]{963400}+1.0}) & ({\color[HTML]{963400}+0.9}) \\\hline

AEL \cite{hu2021semi} & {\color[HTML]{656565} {\small[}NeurIPS'21{]}} &66.1 &68.3 &71.9 & 74.4 & 78.9 & 77.2 & 77.6 & 78.1 \\
AEL \cite{hu2021semi} + MPMC &&\textbf{68.6} &\textbf{70.3}   &\textbf{73.5} & \textbf{76.1} & \textbf{80.1} & \textbf{79.2} & \textbf{79.3} & \textbf{79.2} \\ 
({\color[HTML]{963400}$\Delta$}) &&  ({\color[HTML]{963400}+2.5}) & ({\color[HTML]{963400}+2.0}) & ({\color[HTML]{963400}+1.6}) & ({\color[HTML]{963400}+1.7}) & ({\color[HTML]{963400}+1.2}) & ({\color[HTML]{963400}+2.0}) & ({\color[HTML]{963400}+1.7}) & ({\color[HTML]{963400}+1.1}) \\\hline

U2PL \cite{wang2022semi} & {\color[HTML]{656565} {\small[}CVPR'22{]}} &  68.0 & 69.2 & 73.7 & 76.2 & 79.5 & 77.2 & 79.0 & 79.3 \\
U2PL  \cite{wang2022semi} + MPMC &&\textbf{70.0} & \textbf{70.8} & \textbf{75.3} &\textbf{77.8} &\textbf{80.5} &\textbf{79.2} & \textbf{80.7} &\textbf{80.3} \\
({\color[HTML]{963400}$\Delta$}) &&  ({\color[HTML]{963400}+2.0}) & ({\color[HTML]{963400}+1.6}) & ({\color[HTML]{963400}+1.6}) & ({\color[HTML]{963400}+1.6}) & ({\color[HTML]{963400}+1.0}) & ({\color[HTML]{963400}+2.0}) & ({\color[HTML]{963400}+1.7}) & ({\color[HTML]{963400}+1.0}) \\ \hline

UniMatch \cite{yang2023revisiting} & {\color[HTML]{656565} {\small[}CVPR'23{]}} &  75.2 & 77.2 & 78.8 & 79.9 & 81.2 & 80.9 & 81.9 & 80.4 \\
UniMatch \cite{yang2023revisiting} + MPMC &&  \textbf{77.3} & \textbf{78.6} & \textbf{79.8} & \textbf{80.8} & \textbf{81.7} & \textbf{82.5} & \textbf{83.1} & \textbf{81.5} \\
({\color[HTML]{963400}$\Delta$}) &&  ({\color[HTML]{963400}+2.1}) & ({\color[HTML]{963400}+1.4}) & ({\color[HTML]{963400}+1.0}) & ({\color[HTML]{963400}+0.9}) & ({\color[HTML]{963400}+0.5}) & ({\color[HTML]{963400}+1.6}) & ({\color[HTML]{963400}+1.2}) & ({\color[HTML]{963400}+1.1}) \\\hline
\end{tabular}}
\end{subtable}
\caption{Quantitative results of different semi-supervised segmentation methods on Pascal VOC classic and blender set. We report Mean IoU  under various partition protocols and show the improvements ({\color[HTML]{963400}$\Delta$}) over corresponding baseline. 
}
\label{tab:voc}
\end{table*}
\begin{figure*}
\centering
\includegraphics[width=1\linewidth,height=19\baselineskip]{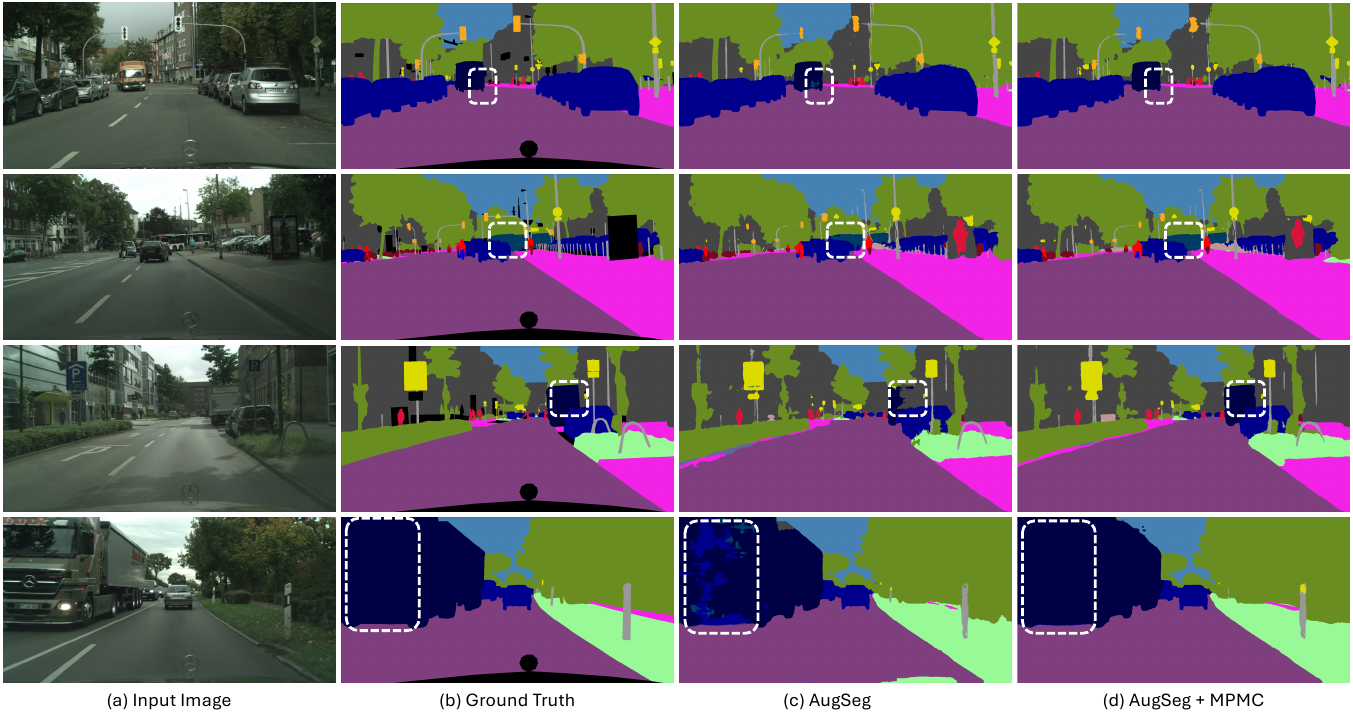}
% \vspace{0.5cm}
\caption{\textbf{Qualitative Results on Cityscapes dataset:} (a) is the original image, (b) is the Ground Truth, (c) are segmentations generated by AugSeg \cite{zhao2023augmentation} compared to (d) which are segmentations generated when our method (MPMC) is integrated to AugSeg. White boxes show the areas where our method improves the baseline.  
}
  \label{fig:unimatchpic}
\end{figure*}

\begin{table*}[]
\begin{subtable}{1\linewidth}
\centering
\resizebox{1\linewidth}{!}{%
\begin{tabular}{cr|cccc|cccc}
\hline
\multicolumn{2}{c|}{\multirow{2}{*}{Method}} & \multicolumn{4}{c|}{\textbf{ResNet-50}} & \multicolumn{4}{c}{\textbf{ResNet-101}} \\ \cline{3-10} 
\multicolumn{2}{c|}{}                        & 1/16   & 1/8   & 1/4   & 1/2   & 1/16   & 1/8   & 1/4   & 1/2   \\ \hline
\textit{Supervised} & & 63.34 & 68.73 & 74.14 & 76.62 & 66.3 & 72.8 & 75.0 & 78.0\\                                               
CPS \cite{chen2021semi} & {\color[HTML]{656565} {\small[}CVPR'21{]}} & 69.79 & 74.39 & 76.85 & 78.64 & 69.8 & 74.3 & 74.6 & 76.8 \\
PS-MT \cite{liu2022perturbed} & {\color[HTML]{656565} {\small[}CVPR'22{]}} &- & 75.76 & 76.92 & 77.64 & - & 76.9 & 77.6 & 79.1 \\
PCR \cite{xu2022semi} & {\color[HTML]{656565} {\small[}NeurIPS'22{]}} &- & - & - &- & 73.4 & 76.3 & 78.4 & 79.1 \\
CFCG  \cite{li2023cfcg} & {\color[HTML]{656565} {\small[}ICCV'23{]}} &76.1 &78.9 &79.3 &80.1 & 77.8 & 79.6 & 80.4 & 80.9 \\ \hline

AugSeg \cite{zhao2023augmentation} & {\color[HTML]{656565} {\small[}CVPR'21{]}} & 73.7 & 76.4 & 78.7 & 79.3 & 75.2 & 77.8 & 79.6 & 80.4 \\
AugSeg \cite{zhao2023augmentation} + MPMC & & \textbf{75.3} & \textbf{77.7} & \textbf{79.8} & \textbf{80.3} & \textbf{76.9} & \textbf{79.2} & \textbf{80.9} & \textbf{81.3} \\
({\color[HTML]{963400}$\Delta$}) & & ({\color[HTML]{963400}+1.6}) & ({\color[HTML]{963400}+1.3}) & ({\color[HTML]{963400}+1.1}) & ({\color[HTML]{963400}+1.0}) & ({\color[HTML]{963400}+1.7}) & ({\color[HTML]{963400}+1.4}) & ({\color[HTML]{963400}+1.3}) & ({\color[HTML]{963400}+0.9}) \\\hline

AEL \cite{hu2021semi} & {\color[HTML]{656565} {\small[}NeurIPS'21{]}} & 68.2 & 72.7 & 74.9 & 77.5 & 74.5 & 75.6 & 77.5 & 79.0 \\
AEL \cite{hu2021semi} + MPMC & &\textbf{70.5} &\textbf{74.5} &\textbf{76.6} &\textbf{79.0} &\textbf{76.6} &\textbf{77.5} & \textbf{78.9} &\textbf{80.3} \\
({\color[HTML]{963400}$\Delta$}) & & ({\color[HTML]{963400}+2.3}) & ({\color[HTML]{963400}+1.8}) & ({\color[HTML]{963400}+1.7}) & ({\color[HTML]{963400}+1.5}) & ({\color[HTML]{963400}+2.1}) & ({\color[HTML]{963400}+1.9}) & ({\color[HTML]{963400}+1.4}) & 
({\color[HTML]{963400}+1.3}) \\\hline

U2PL \cite{wang2022semi} & {\color[HTML]{656565} {\small[}CVPR'22{]}} & 69.0 & 73.0 & 76.3 & 78.6 & 74.9 & 76.5 & 78.5 & 79.1 \\
U2PL \cite{wang2022semi} + MPMC & & \textbf{71.2} & \textbf{74.6} & \textbf{77.8} & \textbf{79.8} & \textbf{77.2} & \textbf{78.2} & \textbf{80.0} & \textbf{80.0} \\
({\color[HTML]{963400}$\Delta$}) & & ({\color[HTML]{963400}+2.2}) & ({\color[HTML]{963400}+1.6}) & ({\color[HTML]{963400}+1.5}) & ({\color[HTML]{963400}+1.2}) & ({\color[HTML]{963400}+2.3}) & ({\color[HTML]{963400}+1.7}) & ({\color[HTML]{963400}+1.5}) & 
({\color[HTML]{963400}+0.9}) \\\hline

UniMatch \cite{yang2023revisiting} & {\color[HTML]{656565} {\small[}CVPR'23{]}} & 75.0 & 76.8 & 77.5 & 78.6 & 76.6 & 77.9 & 79.2 & 79.5\\
UniMatch \cite{yang2023revisiting} + MPMC & &\textbf{76.8} &\textbf{78.0} &\textbf{78.6} &\textbf{79.5} & \textbf{78.5} &\textbf{79.2} & \textbf{80.9} & \textbf{81.3}\\
({\color[HTML]{963400}$\Delta$}) & & ({\color[HTML]{963400}+1.8}) & ({\color[HTML]{963400}+1.2}) & ({\color[HTML]{963400}+1.1}) & ({\color[HTML]{963400}+0.9}) & ({\color[HTML]{963400}+1.9}) & ({\color[HTML]{963400}+1.3}) & ({\color[HTML]{963400}+1.7}) & 
({\color[HTML]{963400}+1.8}) \\\hline

\hline
\end{tabular}}
\end{subtable}
\caption{Quantitative results of different semi-supervised segmentation methods on Cityscapes validation set. We report Mean IoU  under various partition protocols and show the improvements ({\color[HTML]{963400}$\Delta$}) over the corresponding baseline. 
}
\label{tab:ct}
\end{table*}

\begin{table*}[]
\begin{subtable}{1\linewidth}
\centering
\resizebox{0.5\linewidth}{!}{%
\begin{tabular}{cr|cc}
\hline
\multicolumn{2}{c|}{Method} & 3 cases & 7 cases \\ \hline

\textit{Supervised} & & 41.5 & 62.5 \\
UA-MT \cite{yu2019uncertainty} & {\color[HTML]{656565} {\small[}MICCAI'19{]}} & 61.0 & 81.5 \\
CPS \cite{chen2021semi} & {\color[HTML]{656565} {\small[}CVPR'21{]}} & 60.3 & 83.3 \\
CNN \& Trans \cite{luo2022semi} & {\color[HTML]{656565} {\small[}PMLR'22{]}} & 65.6 & 86.4 \\\hline

AugSeg \cite{zhao2023augmentation} & {\color[HTML]{656565} {\small[}CVPR'21{]}} & 85.1 & 87.3 \\ 
AugSeg \cite{zhao2023augmentation} + MPMC & & \textbf{88.6} & \textbf{88.4} \\
({\color[HTML]{963400}$\Delta$}) & & ({\color[HTML]{963400}+3.5}) & ({\color[HTML]{963400}+1.1}) \\\hline

AEL \cite{hu2021semi} & {\color[HTML]{656565} {\small[}NeurIPS'21{]}} & 82.7 & 87.6 \\
AEL \cite{hu2021semi} + MPMC & & \textbf{85.4} & \textbf{88.5} \\
({\color[HTML]{963400}$\Delta$}) & & ({\color[HTML]{963400}+2.7}) & ({\color[HTML]{963400}+0.9}) \\\hline

U2PL \cite{wang2022semi} & {\color[HTML]{656565} {\small[}CVPR'22{]}} & 84.2 & 89.2 \\
U2PL \cite{wang2022semi} + MPMC & & \textbf{87.4} & \textbf{90.5} \\
({\color[HTML]{963400}$\Delta$}) & & ({\color[HTML]{963400}+3.2}) & ({\color[HTML]{963400}+1.3}) \\\hline

UniMatch \cite{yang2023revisiting} & {\color[HTML]{656565} {\small[}CVPR'23{]}} & 88.9 & 89.9 \\
UniMatch \cite{yang2023revisiting} + MPMC & & \textbf{91.8} & \textbf{90.4} \\
({\color[HTML]{963400}$\Delta$}) & & ({\color[HTML]{963400}+2.9}) & ({\color[HTML]{963400}+0.5}) \\\hline

\hline
\end{tabular}
}
\end{subtable}
\caption{Quantitative results of different semi-supervised segmentation methods on ACDC dataset with 3/7 labeled cases. We report Dice Similarity Coefficient (DSC) metric averaged over 3 classes. We use ResNet-101 for all experiments in this dataset.
}
\label{tab:medical}
\end{table*}
\noindent
\textbf{Qualitative Results}
In Fig. \ref{fig:unimatchpic} we visualise the results of integration of our method (MPMC) to AugSeg \cite{zhao2023augmentation}. Introducing patch-level supervision with the integration of MPMC, leads the SSS model to better segment confusing classes. Among comparisons we show that MPMC leads to better segmentation of pixels regions of confusing classes like bus,  train and truck. These classes are often misclassified not only because of sharing visual similarity but also having low training instances.

% \begin{figure*}
% \centering
% %\includegraphics[width=1\linewidth,height=19\baselineskip]{pics/picaugseg.pdf}
% \includegraphics[width=1\linewidth,height=19\baselineskip]{pics/paper_augseg_images.pdf}
% % \vspace{0.5 cm}
% % \vspace{0.1ex}
% \caption{\textbf{Qualitative Results on Cityscapes dataset:} (a) is the original image, (b) is the Ground Truth, (c) are segmentations generated by AugSeg \cite{zhao2023augmentation} compared to (d) which are segmentations generated when our method (MPMC) is integrated to AugSeg. The dashed white boxes show the areas where our method improves the baseline. \cite{zhao2023augmentation}  
% }
%   \label{fig:augsegpic}
% \end{figure*}

\subsection{Ablation Studies}
\begin{figure}[ht!]
    \centering
   \includegraphics[scale=0.4]{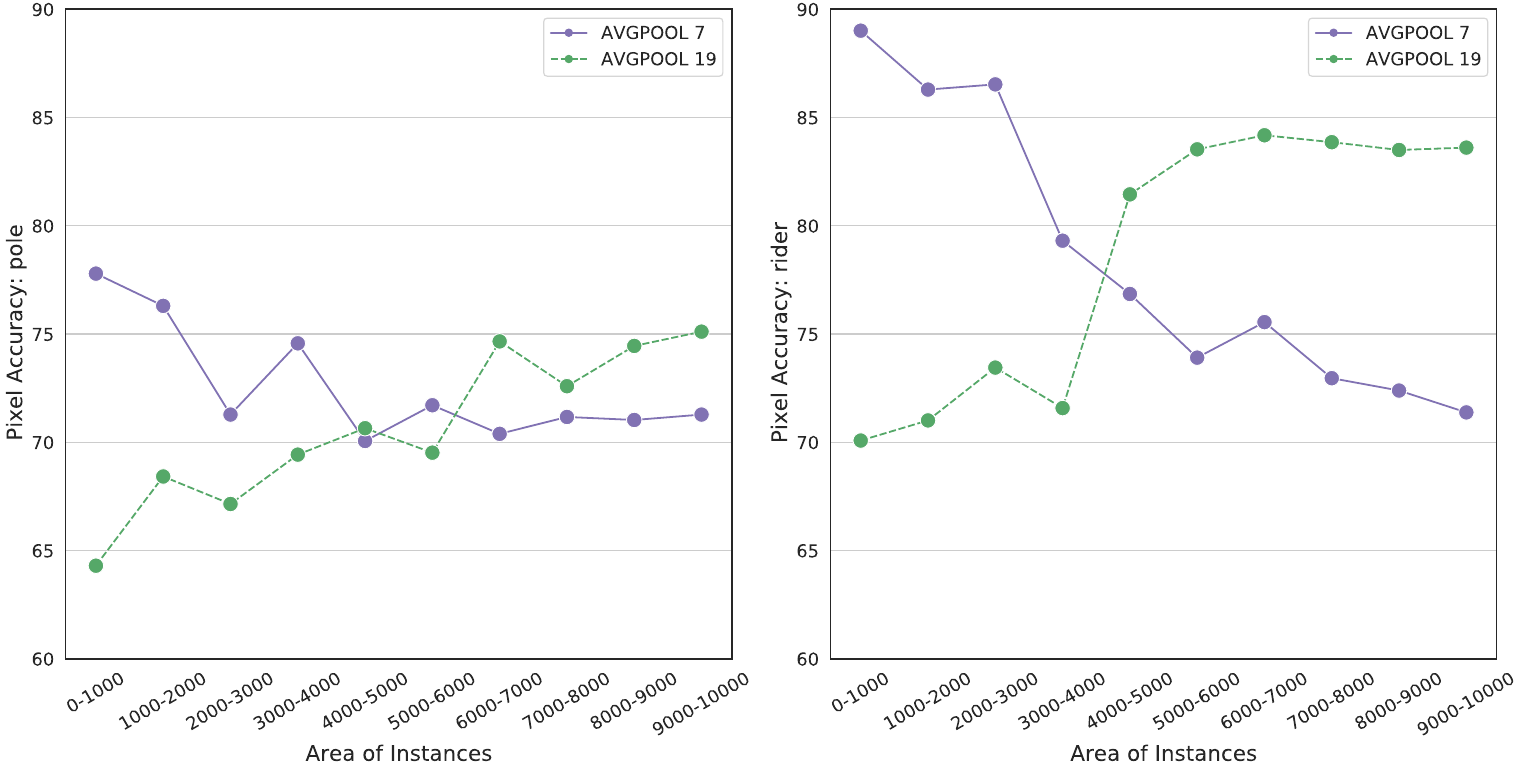} 
   \caption{\textbf{ Pixel accuracy across different scales with respect to  instance size.} For two different scales, we compare the pixel accuracy with respect to size of instances for two random classes: pole and rider.
 }
  \label{fig:barplot}
\end{figure}

We conduct experiments to study the impact of various components of our approach. All  experiments in this section were conducted on $\frac{1}{16}$ data partition of the Cityscapes dataset, and use AugSeg  \cite{zhao2023augmentation} as the baseline.\\
\textbf{Segmentation performance at different scales.}
Here we analyse how scaling of input features for patch-classification, improves the performance of the SSS method. We consider two scales - average pool kernels $7 \times 7$ and $19 \times 19 $. Fig. \ref{fig:barplot} illustrates for two random classes pole and rider, that average pooling with a kernel size of $7 \times 7$  is better at enhancing the pixel accuracy for instances with smaller areas, whereas the $19 \times 19$ kernel size is better at improving accuracy for instances with larger areas. This can be attributed to the larger receptive field of the $19 \times 19$ kernel, enabling more effective contextual learning for larger regions. While the smaller receptive field of $7 \times 7$ kernel is better able to discern smaller objects.\\
\textbf{Analysing class features}
\begin{figure}[ht!]
    \centering
   \includegraphics[scale =0.70]{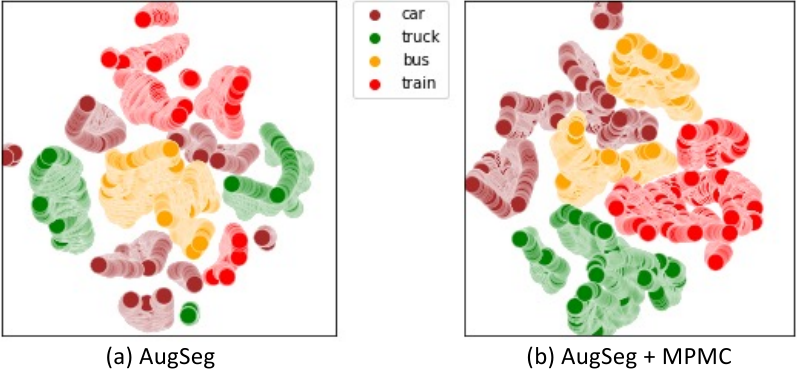}
   \caption{\textbf{ Analysis of class features } (a) t-SNE of features predicted by AugSeg \cite{zhao2023augmentation}, (b) t-SNE of features predicted by AugSeg \cite{zhao2023augmentation} + MPMC. 
 }
  \label{fig:tsne}
\end{figure}
The crux of our method is to better discriminate pixel regions between different classes. Here we analyze if MPMC integrated to AugSeg \cite{zhao2023augmentation} can better  discriminate features than the baseline for different classes in Cityscapes Dataset. We use the pixel features from the penultimate layer of the segmentation network to analyse via t-SNE. Fig \ref{fig:tsne} illustrates for four random classes - car, bus, train and truck, integration of MPMC leads to better discrimination between the classes, \textit{i.e.}, compared to AugSeg \cite{zhao2023augmentation}, the features of \textit{train} and \textit{truck} are more clustered and separated from the rest.\\
\textbf{Influence of MPMC on patch-level segmentation.}
\begin{table}[]
\begin{subtable}{1\linewidth}
\centering
\resizebox{1\linewidth}{!}{%
\begin{tabular}{@{}c|cccccccccc@{}}
\toprule
\textbf{1 - hamming loss} & 90+   & 80-90 & 70-80 & 60-70 & 50-60 & 40-50 & 30-40 & 20-30 & 10-20 & 0-10  \\ \midrule
\textbf{\% patches}      & 12.6  & 25.2  & 20.5  & 8.4   & 5.1   & 5.2   & 4.1   & 10.2  & 2.4   & 6.3   \\\midrule
\textbf{Avg MIoU} {\color[HTML]{963400}$\Delta$}    & {\color[HTML]{963400}+4.05} & {\color[HTML]{963400}+1.57} & {\color[HTML]{963400}+0.88} & {\color[HTML]{963400}+1.55} & {\color[HTML]{963400}+0.91} & {\color[HTML]{963400}+1.15} & {\color[HTML]{963400}+0.49} & {\color[HTML]{963400}+0.10} & {\color[HTML]{963400}-1.90} & {\color[HTML]{963400}-2.20} \\ \bottomrule
\end{tabular}}
\end{subtable}
\caption{Analysis of the influence of MPMC on segmentation at patch-level.}
\label{tab:mlcperfabl}
\end{table}
Our method MPMC improves segmentation by introducing patch classification as an auxiliary task. We analyse at patch level the correlation between multi-label accuracy and the Mean IoU performance of the segmentation network. We use (1 - hamming loss) \cite{tsoumakas2008multi} as a measure of multi-label classification and  use the basic teacher-student framework mentioned in Section \ref{Preliminary} as our baseline. Table \ref{tab:mlcperfabl} illustrates that the avg MIoU improvement is highest ($4.05\%$) in patches having more than $90\%$ multi-label classification accuracy showing that MPMC improves segmentation at the patch level.\\ 
\textbf{Impact of MPMC on semantic segmentation in limited labeled data scenarios.}
\begin{table}
\parbox{.45\linewidth}{
\centering
\begin{tabular}{c|ccc}
\hline
                   & \multicolumn{3}{c}{\# Labeled Data}                                                     \\ \cline{2-4} 
\multirow{-2}{*}{} & 186                         & 372                         & 744                         \\ \hline
w/o MPMC            & 66.3                        & 72.8                        & 75.0                        \\
w/ MPMC             & \textbf{68.2}               & \textbf{74.3}               & \textbf{76.3}               \\
Gain/{\color[HTML]{963400}$\Delta$}         & {\color[HTML]{963400} +1.9} & {\color[HTML]{963400} +1.5} & {\color[HTML]{963400} +1.3} \\ \hline
\end{tabular} 
\caption{Impact of MPMC on segmentation (MIoU) when trained with different volumes of labeled data. MPMC shows the highest improvement (1.9\%) in the least labeled data partition.}
\label{tab:mlclimdataabl}
% \caption{Analysis of the impact of MPMC on segmentation (MIoU) when trained with different quantities of labeled data. We observe that MPMC provides the highest improvement in segmentation performance in the partition with the least amount of labeled data}
}
\hfill
\parbox{.5\linewidth}{
\centering
\begin{tabular}{cc|ccc}
\hline
MPMCL & MPMCUL & \begin{tabular}[c]{@{}c@{}}1/16\\ (186)\end{tabular} & \begin{tabular}[c]{@{}c@{}}1/8\\ (372)\end{tabular} & \begin{tabular}[c]{@{}c@{}}1/4\\ (744)\end{tabular} \\ \hline
& & 68.4 & 73.5 & 74.2 \\
\checkmark & & 70.8 & 75.4 & 75.7 \\
& \checkmark & 71.3 & 76.0 & 76.6 \\
\checkmark & \checkmark & \textbf{73.1} & \textbf{76.7} & \textbf{77.3} \\ \hline
\end{tabular} 
\caption{Ablation study of different component: MPMC on labeled images (MPMCL) and  MPMC on unlabeled images (MPMCUL)}
\label{tab:diffcompabl}
}
\end{table}
\noindent
In Table \ref{tab:mlclimdataabl} shows the impact of MPMC on segmentation when trained with only labeled data.
 We use the basic teacher-student framework mentioned in Section \ref{Preliminary} as our baseline, which achieves MIoU of $66.3$, $72.8$ and $75.0$ in  $186$, $372$ and $744$ partitions of labeled images respectively. After introducing PMC, there is an improvement of $1.9\%$, $1.5\%$ and $1.3\%$ in  $186$, $372$ and $744$ partitions of labeled images respectively. We observe that MPMC improves segmentation performance in limited-labeled data scenarios, with the improvement highest for least labeled data, i.e., $1.9\%$ for $186$ labeled images.\\
\textbf{Analysis of different components}
% \begin{table}[hbt!]
% \centering
% \begin{tabular}{cc|ccc}
% \hline
% PMCL & PMCUL & \begin{tabular}[c]{@{}c@{}}1/16\\ (186)\end{tabular} & \begin{tabular}[c]{@{}c@{}}1/8\\ (372)\end{tabular} & \begin{tabular}[c]{@{}c@{}}1/4\\ (744)\end{tabular} \\ \hline
%      &       & 68.4                                                 & 73.5                                                & 74.2                                                \\
% \checkmark    &       & 70.8                                                 & 75.4                                                & 75.7                                                \\
%      & \checkmark     & 71.3                                                 & 76.0                                                & 76.6                                                \\
% \checkmark    & \checkmark     & 73.1                                                 & 76.7                                                & 77.3                                                \\ \hline
% \end{tabular}
% \caption{Ablation study of different component: MPMC on labeled images (PMCL) and  MPMC on unlabeled images (PMCUL)}
% \label{tab:diffcompabl}
% \end{table}
We ablate each component of MPMC with basic teacher-student framework (Section \ref{Preliminary}) as the baseline. MPMC on labeled images (MPMCL), which introduces multi-scale patch-based classification on only the labeled data, achieves an improvement of $2.4\%$, $1.9\%$, and $1.5\%$ under $\frac{1}{16}$, $\frac{1}{8}$ and $\frac{1}{4}$ partition protocols respectively. MPMC on unlabeled images (MPMCUL) leads to a further  improvement of  $2.9\%$, $2.5\%$, and $2.4\%$ under $\frac{1}{16}$, $\frac{1}{8}$ and $\frac{1}{4}$ partition protocols respectively. Thus, MPMC is effective on both labeled and unlabeled instances in SSS pipelines.

\section{Conclusion}
In this paper, we propose a novel Multi-scale Patch-based Multi-label Classifier (MPMC), which introduces patch-level context information in the form of an auxiliary task of multi-scale patch-based multi-label classification. When jointly trained with a SSS model, the patch classifier provides patch-level labels, identifying the classes present within a region. This facilitates the  elimination of distractors and enhances the classification of small object segments by the baseline method. Further, the MPMC predictions are used to learn an adaptive weight that reduces the influence of noisy pseudo-labels in the teacher-student framework. We perform thorough analysis of the importance of multi-scale patch-level contextual information in improving segmentation especially in low data scenarios. Further, we demonstrate the efficacy of our proposed parameterized module MPMC by integrating it into four SSS methodologies and showing consistent improvements in  Cityscapes, Pascal VOC and ACDC benchmarks.

\bibliographystyle{splncs04}
\bibliography{main}

\begin{thebibliography}{10}
\providecommand{\url}[1]{\texttt{#1}}
\providecommand{\urlprefix}{URL }
\providecommand{\doi}[1]{https://doi.org/#1}

\bibitem{alonso2021semi}
Alonso, I., Sabater, A., Ferstl, D., Montesano, L., Murillo, A.C.: Semi-supervised semantic segmentation with pixel-level contrastive learning from a class-wise memory bank. In: Proceedings of the IEEE/CVF International Conference on Computer Vision. pp. 8219--8228 (2021)

\bibitem{bachman2014learning}
Bachman, P., Alsharif, O., Precup, D.: Learning with pseudo-ensembles. Advances in neural information processing systems  \textbf{27} (2014)

\bibitem{bernard2018deep}
Bernard, O., Lalande, A., Zotti, C., Cervenansky, F., Yang, X., Heng, P.A., Cetin, I., Lekadir, K., Camara, O., Ballester, M.A.G., et~al.: Deep learning techniques for automatic mri cardiac multi-structures segmentation and diagnosis: is the problem solved? IEEE transactions on medical imaging  \textbf{37}(11),  2514--2525 (2018)

\bibitem{berthelot2019mixmatch}
Berthelot, D., Carlini, N., Goodfellow, I., Papernot, N., Oliver, A., Raffel, C.A.: Mixmatch: A holistic approach to semi-supervised learning. Advances in neural information processing systems  \textbf{32} (2019)

\bibitem{chen2021semisupervised}
Chen, H., Jin, Y., Jin, G., Zhu, C., Chen, E.: Semisupervised semantic segmentation by improving prediction confidence. IEEE Transactions on Neural Networks and Learning Systems  (2021)

\bibitem{chen2018encoder}
Chen, L.C., Zhu, Y., Papandreou, G., Schroff, F., Adam, H.: Encoder-decoder with atrous separable convolution for semantic image segmentation. In: Proceedings of the European conference on computer vision (ECCV). pp. 801--818 (2018)

\bibitem{chen2021semi}
Chen, X., Yuan, Y., Zeng, G., Wang, J.: Semi-supervised semantic segmentation with cross pseudo supervision. In: Proceedings of the IEEE/CVF Conference on Computer Vision and Pattern Recognition. pp. 2613--2622 (2021)

\bibitem{cordts2016cityscapes}
Cordts, M., Omran, M., Ramos, S., Rehfeld, T., Enzweiler, M., Benenson, R., Franke, U., Roth, S., Schiele, B.: The cityscapes dataset for semantic urban scene understanding. In: Proceedings of the IEEE conference on computer vision and pattern recognition. pp. 3213--3223 (2016)

\bibitem{devries2017improved}
DeVries, T., Taylor, G.W.: Improved regularization of convolutional neural networks with cutout. arXiv preprint arXiv:1708.04552  (2017)

\bibitem{durasov2024zigzag}
Durasov, N., Dorndorf, N., Le, H., Fua, P.: Zigzag: Universal sampling-free uncertainty estimation through two-step inference. Transactions on Machine Learning Research  (2024)

\bibitem{durasov2024enabling}
Durasov, N., Oner, D., Donier, J., Le, H., Fua, P.: Enabling uncertainty estimation in iterative neural networks. In: Forty-first International Conference on Machine Learning (2024)

\bibitem{everingham2010pascal}
Everingham, M., Van~Gool, L., Williams, C.K., Winn, J., Zisserman, A.: The pascal visual object classes (voc) challenge. International journal of computer vision  \textbf{88}(2),  303--338 (2010)

\bibitem{fan2022ucc}
Fan, J., Gao, B., Jin, H., Jiang, L.: Ucc: Uncertainty guided cross-head co-training for semi-supervised semantic segmentation. In: Proceedings of the IEEE/CVF Conference on Computer Vision and Pattern Recognition. pp. 9947--9956 (2022)

\bibitem{french2019semi}
French, G., Laine, S., Aila, T., Mackiewicz, M., Finlayson, G.: Semi-supervised semantic segmentation needs strong, varied perturbations. arXiv preprint arXiv:1906.01916  (2019)

\bibitem{ge2018multi}
Ge, W., Yang, S., Yu, Y.: Multi-evidence filtering and fusion for multi-label classification, object detection and semantic segmentation based on weakly supervised learning. In: Proceedings of the IEEE conference on computer vision and pattern recognition. pp. 1277--1286 (2018)

\bibitem{grandvalet2004semi}
Grandvalet, Y., Bengio, Y.: Semi-supervised learning by entropy minimization. Advances in neural information processing systems  \textbf{17} (2004)

\bibitem{guo2021long}
Guo, H., Wang, S.: Long-tailed multi-label visual recognition by collaborative training on uniform and re-balanced samplings. In: Proceedings of the IEEE/CVF Conference on Computer Vision and Pattern Recognition. pp. 15089--15098 (2021)

\bibitem{hariharan2011semantic}
Hariharan, B., Arbel{\'a}ez, P., Bourdev, L., Maji, S., Malik, J.: Semantic contours from inverse detectors. In: 2011 international conference on computer vision. pp. 991--998. IEEE (2011)

\bibitem{he2016deep}
He, K., Zhang, X., Ren, S., Sun, J.: Deep residual learning for image recognition. In: Proceedings of the IEEE conference on computer vision and pattern recognition. pp. 770--778 (2016)

\bibitem{hu2021semi}
Hu, H., Wei, F., Hu, H., Ye, Q., Cui, J., Wang, L.: Semi-supervised semantic segmentation via adaptive equalization learning. Advances in Neural Information Processing Systems  \textbf{34},  22106--22118 (2021)

\bibitem{huang2015learning}
Huang, J., Li, G., Huang, Q., Wu, X.: Learning label specific features for multi-label classification. In: 2015 IEEE International Conference on Data Mining. pp. 181--190. IEEE (2015)

\bibitem{huo2021atso}
Huo, X., Xie, L., He, J., Yang, Z., Zhou, W., Li, H., Tian, Q.: Atso: Asynchronous teacher-student optimization for semi-supervised image segmentation. In: Proceedings of the IEEE/CVF conference on computer vision and pattern recognition. pp. 1235--1244 (2021)

\bibitem{ke2020guided}
Ke, Z., Qiu, D., Li, K., Yan, Q., Lau, R.W.: Guided collaborative training for pixel-wise semi-supervised learning. In: European conference on computer vision. pp. 429--445. Springer (2020)

\bibitem{Le_CVPRW19}
Le, H., Goncalves, B., Samaras, D., Lynch, H.: Weakly labeling the antarctic: The penguin colony case. In: CVPRW (June 2019)

\bibitem{Le_RS22}
Le, H., Samaras, D., Lynch, H.J.: A convolutional neural network architecture designed for the automated survey of seabird colonies. Remote Sensing in Ecology and Conservation  \textbf{8}(2),  251--262 (2022)

\bibitem{Le_ECCV18}
Le, H., Vicente, T.F.Y., Nguyen, V., Hoai, M., Samaras, D.: {A+D Net}: Training a shadow detector with adversarial shadow attenuation. In: European Conference on Computer Vision(ECCV) (2018)

\bibitem{Le_ICCV2017}
Le, H., Yu, C.P., Zelinsky, G., Samaras, D.: Co-localization with category-consistent features and geodesic distance propagation. In: ICCVW (2017)

\bibitem{lee2013pseudo}
Lee, D.H., et~al.: Pseudo-label: The simple and efficient semi-supervised learning method for deep neural networks. In: Workshop on challenges in representation learning, ICML. vol.~3, p.~896 (2013)

\bibitem{li2023cfcg}
Li, S., He, Y., Zhang, W., Zhang, W., Tan, X., Han, J., Ding, E., Wang, J.: Cfcg: Semi-supervised semantic segmentation via cross-fusion and contour guidance supervision. In: Proceedings of the IEEE/CVF International Conference on Computer Vision. pp. 16348--16358 (2023)

\bibitem{10.1609/aaai.v37i2.25244}
Lin, D.: Probability guided loss for long-tailed multi-label image classification. In: Proceedings of the Thirty-Seventh AAAI Conference on Artificial Intelligence and Thirty-Fifth Conference on Innovative Applications of Artificial Intelligence and Thirteenth Symposium on Educational Advances in Artificial Intelligence. AAAI'23/IAAI'23/EAAI'23, AAAI Press (2023). \doi{10.1609/aaai.v37i2.25244}, \url{https://doi.org/10.1609/aaai.v37i2.25244}

\bibitem{lin2016efficient}
Lin, G., Shen, C., Van Den~Hengel, A., Reid, I.: Efficient piecewise training of deep structured models for semantic segmentation. In: Proceedings of the IEEE conference on computer vision and pattern recognition. pp. 3194--3203 (2016)

\bibitem{liu2015parsenet}
Liu, W., Rabinovich, A., Berg, A.C.: Parsenet: Looking wider to see better. arXiv preprint arXiv:1506.04579  (2015)

\bibitem{liu2020energy}
Liu, W., Wang, X., Owens, J., Li, Y.: Energy-based out-of-distribution detection. Advances in neural information processing systems  \textbf{33},  21464--21475 (2020)

\bibitem{liu2022perturbed}
Liu, Y., Tian, Y., Chen, Y., Liu, F., Belagiannis, V., Carneiro, G.: Perturbed and strict mean teachers for semi-supervised semantic segmentation. In: Proceedings of the IEEE/CVF Conference on Computer Vision and Pattern Recognition. pp. 4258--4267 (2022)

\bibitem{luo2022semi}
Luo, X., Hu, M., Song, T., Wang, G., Zhang, S.: Semi-supervised medical image segmentation via cross teaching between cnn and transformer. In: International Conference on Medical Imaging with Deep Learning. pp. 820--833. PMLR (2022)

\bibitem{mcclosky2006effective}
McClosky, D., Charniak, E., Johnson, M.: Effective self-training for parsing. In: Proceedings of the Human Language Technology Conference of the NAACL, Main Conference. pp. 152--159 (2006)

\bibitem{oliver2018realistic}
Oliver, A., Odena, A., Raffel, C.A., Cubuk, E.D., Goodfellow, I.: Realistic evaluation of deep semi-supervised learning algorithms. Advances in neural information processing systems  \textbf{31} (2018)

\bibitem{olsson2021classmix}
Olsson, V., Tranheden, W., Pinto, J., Svensson, L.: Classmix: Segmentation-based data augmentation for semi-supervised learning. In: Proceedings of the IEEE/CVF Winter Conference on Applications of Computer Vision. pp. 1369--1378 (2021)

\bibitem{ouali2020semi}
Ouali, Y., Hudelot, C., Tami, M.: Semi-supervised semantic segmentation with cross-consistency training. In: Proceedings of the IEEE/CVF Conference on Computer Vision and Pattern Recognition. pp. 12674--12684 (2020)

\bibitem{read2011classifier}
Read, J., Pfahringer, B., Holmes, G., Frank, E.: Classifier chains for multi-label classification. Machine learning  \textbf{85},  333--359 (2011)

\bibitem{reiss2021every}
Rei{\ss}, S., Seibold, C., Freytag, A., Rodner, E., Stiefelhagen, R.: Every annotation counts: Multi-label deep supervision for medical image segmentation. In: Proceedings of the IEEE/CVF conference on computer vision and pattern recognition. pp. 9532--9542 (2021)

\bibitem{ridnik2021asymmetric}
Ridnik, T., Ben-Baruch, E., Zamir, N., Noy, A., Friedman, I., Protter, M., Zelnik-Manor, L.: Asymmetric loss for multi-label classification. In: Proceedings of the IEEE/CVF International Conference on Computer Vision. pp. 82--91 (2021)

\bibitem{rizve2021defense}
Rizve, M.N., Duarte, K., Rawat, Y.S., Shah, M.: In defense of pseudo-labeling: An uncertainty-aware pseudo-label selection framework for semi-supervised learning. arXiv preprint arXiv:2101.06329  (2021)

\bibitem{sajjadi2016regularization}
Sajjadi, M., Javanmardi, M., Tasdizen, T.: Regularization with stochastic transformations and perturbations for deep semi-supervised learning. Advances in neural information processing systems  \textbf{29} (2016)

\bibitem{selvaraju2017grad}
Selvaraju, R.R., Cogswell, M., Das, A., Vedantam, R., Parikh, D., Batra, D.: Grad-cam: Visual explanations from deep networks via gradient-based localization. In: Proceedings of the IEEE international conference on computer vision. pp. 618--626 (2017)

\bibitem{shi2018transductive}
Shi, W., Gong, Y., Ding, C., Tao, Z.M., Zheng, N.: Transductive semi-supervised deep learning using min-max features. In: Proceedings of the European Conference on Computer Vision (ECCV). pp. 299--315 (2018)

\bibitem{sohn2020fixmatch}
Sohn, K., Berthelot, D., Carlini, N., Zhang, Z., Zhang, H., Raffel, C.A., Cubuk, E.D., Kurakin, A., Li, C.L.: Fixmatch: Simplifying semi-supervised learning with consistency and confidence. Advances in neural information processing systems  \textbf{33},  596--608 (2020)

\bibitem{strudel2021segmenter}
Strudel, R., Garcia, R., Laptev, I., Schmid, C.: Segmenter: Transformer for semantic segmentation. In: Proceedings of the IEEE/CVF international conference on computer vision. pp. 7262--7272 (2021)

\bibitem{sun2023daw}
Sun, R., Mai, H., Zhang, T., Wu, F.: {DAW}: Exploring the better weighting function for semi-supervised semantic segmentation. In: Thirty-seventh Conference on Neural Information Processing Systems (2023), \url{https://openreview.net/forum?id=KRlG7NJUCD}

\bibitem{tsoumakas2008multi}
Tsoumakas, G., Katakis, I.: Multi-label classification. Data Warehousing and Mining: Concepts, Methodologies, Tools, and Applications: Concepts, Methodologies, Tools, and Applications  \textbf{3}, ~64 (2008)

\bibitem{verma2019interpolation}
Verma, V., Kawaguchi, K., Lamb, A., Kannala, J., Bengio, Y., Lopez-Paz, D.: Interpolation consistency training for semi-supervised learning. arXiv preprint arXiv:1903.03825  (2019)

\bibitem{wang2021can}
Wang, H., Liu, W., Bocchieri, A., Li, Y.: Can multi-label classification networks know what they don’t know? Advances in Neural Information Processing Systems  \textbf{34},  29074--29087 (2021)

\bibitem{wang2022semi}
Wang, Y., Wang, H., Shen, Y., Fei, J., Li, W., Jin, G., Wu, L., Zhao, R., Le, X.: Semi-supervised semantic segmentation using unreliable pseudo-labels. In: Proceedings of the IEEE/CVF Conference on Computer Vision and Pattern Recognition. pp. 4248--4257 (2022)

\bibitem{wu2020distribution}
Wu, T., Huang, Q., Liu, Z., Wang, Y., Lin, D.: Distribution-balanced loss for multi-label classification in long-tailed datasets. In: Computer Vision--ECCV 2020: 16th European Conference, Glasgow, UK, August 23--28, 2020, Proceedings, Part IV 16. pp. 162--178. Springer (2020)

\bibitem{xu2022semi}
Xu, H.M., Liu, L., Bian, Q., Yang, Z.: Semi-supervised semantic segmentation with prototype-based consistency regularization. arXiv preprint arXiv:2210.04388  (2022)

\bibitem{yang2023revisiting}
Yang, L., Qi, L., Feng, L., Zhang, W., Shi, Y.: Revisiting weak-to-strong consistency in semi-supervised semantic segmentation. In: Proceedings of the IEEE/CVF Conference on Computer Vision and Pattern Recognition. pp. 7236--7246 (2023)

\bibitem{yang2022st++}
Yang, L., Zhuo, W., Qi, L., Shi, Y., Gao, Y.: St++: Make self-training work better for semi-supervised semantic segmentation. In: Proceedings of the IEEE/CVF Conference on Computer Vision and Pattern Recognition. pp. 4268--4277 (2022)

\bibitem{yarowsky1995unsupervised}
Yarowsky, D.: Unsupervised word sense disambiguation rivaling supervised methods. In: 33rd annual meeting of the association for computational linguistics. pp. 189--196 (1995)

\bibitem{yu2019uncertainty}
Yu, L., Wang, S., Li, X., Fu, C.W., Heng, P.A.: Uncertainty-aware self-ensembling model for semi-supervised 3d left atrium segmentation. In: International Conference on Medical Image Computing and Computer-Assisted Intervention. pp. 605--613. Springer (2019)

\bibitem{yu2014multi}
Yu, Y., Pedrycz, W., Miao, D.: Multi-label classification by exploiting label correlations. Expert Systems with Applications  \textbf{41}(6),  2989--3004 (2014)

\bibitem{yuan2021simple}
Yuan, J., Liu, Y., Shen, C., Wang, Z., Li, H.: A simple baseline for semi-supervised semantic segmentation with strong data augmentation. In: Proceedings of the IEEE/CVF International Conference on Computer Vision. pp. 8229--8238 (2021)

\bibitem{yun2019cutmix}
Yun, S., Han, D., Oh, S.J., Chun, S., Choe, J., Yoo, Y.: Cutmix: Regularization strategy to train strong classifiers with localizable features. In: Proceedings of the IEEE/CVF international conference on computer vision. pp. 6023--6032 (2019)

\bibitem{zhang2017mixup}
Zhang, H., Cisse, M., Dauphin, Y.N., Lopez-Paz, D.: mixup: Beyond empirical risk minimization. arXiv preprint arXiv:1710.09412  (2017)

\bibitem{zhao2023augmentation}
Zhao, Z., Yang, L., Long, S., Pi, J., Zhou, L., Wang, J.: Augmentation matters: A simple-yet-effective approach to semi-supervised semantic segmentation. In: Proceedings of the IEEE/CVF Conference on Computer Vision and Pattern Recognition. pp. 11350--11359 (2023)

\bibitem{zhong2021pixel}
Zhong, Y., Yuan, B., Wu, H., Yuan, Z., Peng, J., Wang, Y.X.: Pixel contrastive-consistent semi-supervised semantic segmentation. In: Proceedings of the IEEE/CVF International Conference on Computer Vision. pp. 7273--7282 (2021)

\bibitem{zhou2020time}
Zhou, T., Wang, S., Bilmes, J.: Time-consistent self-supervision for semi-supervised learning. In: International Conference on Machine Learning. pp. 11523--11533. PMLR (2020)

\bibitem{zhou2019context}
Zhou, Y., Sun, X., Zha, Z.J., Zeng, W.: Context-reinforced semantic segmentation. In: Proceedings of the IEEE/CVF Conference on Computer Vision and Pattern Recognition. pp. 4046--4055 (2019)

\bibitem{zou2020pseudoseg}
Zou, Y., Zhang, Z., Zhang, H., Li, C.L., Bian, X., Huang, J.B., Pfister, T.: Pseudoseg: Designing pseudo labels for semantic segmentation. arXiv preprint arXiv:2010.09713  (2020)

\bibitem{zuo2021self}
Zuo, S., Yu, Y., Liang, C., Jiang, H., Er, S., Zhang, C., Zhao, T., Zha, H.: Self-training with differentiable teacher. arXiv preprint arXiv:2109.07049  (2021)

\end{thebibliography}
%\import{Analysing multi-scale design of MPMC}{supplementary.tex}
\newpage
\appendix
\title{Beyond Pixels: Semi-Supervised Semantic Segmentation  with a Multi-scale Patch-based Multi-Label Classifier \\ \textit{Supplementary Material}} 

% TODO REVIEW: If the paper title is too long for the running head, you can set
% an abbreviated paper title here. If not, comment out.
\titlerunning{Beyond Pixels}

% % TODO FINAL: Replace with your author list. 
% % Include the authors' OCRID for the camera-ready version, if at all possible.
\author{  }
\institute{ }
% Second Author\inst{2,3}\orcidlink{1111-2222-3333-4444} \and
% Third Author\inst{3}\orcidlink{2222--3333-4444-5555}}

% % TODO FINAL: Replace with an abbreviated list of authors.
\authorrunning{P.Howlader et al.}
% % First names are abbreviated in the running head.
% % If there are more than two authors, 'et al.' is used.

% % TODO FINAL: Replace with your institution list.
% \institute{Princeton University, Princeton NJ 08544, USA \and
% Springer Heidelberg, Tiergartenstr.~17, 69121 Heidelberg, Germany
% \email{lncs@springer.com}\\
% \url{http://www.springer.com/gp/computer-science/lncs} \and
% ABC Institute, Rupert-Karls-University Heidelberg, Heidelberg, Germany\\
% \email{\{abc,lncs\}@uni-heidelberg.de}}

\maketitle

\textbf{Summary:}
In this supplementary material, we provide additional analysis and results of our proposed SSS framework: \textbf{MPMC}, including:
\begin{itemize}
    \item Analysing multi-scale design of MPMC 
    \item Analysis of training hyper-parameters
    \item Analysis of per-class performance
    \item Relationship between multi-label classifier confidence and noisy pseudo-labels
    \item Ablation study on the plug-in position of classifier
    \item Performance gains with Vision Transformer
    \item Qualitative Results
\end{itemize}
\textbf{Analysing multi-scale design of MPMC.}
We use three scaled features from the first layer feature of the encoder: original feature, average pool of $7 \times 7$ and $19 \times 19$. Results are shown in Table \ref{tab:ablscale}. We observe that each part contributes to the final performance suggesting that different scales can encapsulate complimentary contextual information pertaining to the objects in the scene.
Consequently, this experimental analysis shows the importance of the multi-scale design of our proposed MPMC. 
\begin{table*}[]
\begin{subtable}{1\linewidth}
\centering
\resizebox{0.4\linewidth}{!}{%
\begin{tabular}{ccc|c}
\hline
\begin{tabular}[c]{@{}c@{}}Original\\ Feature\end{tabular}  & \begin{tabular}[c]{@{}c@{}}$7 \times 7$\\ Avgpool\end{tabular} & \begin{tabular}[c]{@{}c@{}}$19 \times 19$\\ Avgpool\end{tabular} & mIoU \\ \hline
     &             &               & 73.7\\
\checkmark      &             &               &74.1\\
     & \checkmark  &                &74.3      \\
     & &\checkmark                &74.1      \\
 \checkmark    & \checkmark  &  & 74.5     \\
 \checkmark    &  &\checkmark &74.4  \\
     & \checkmark &\checkmark   & 74.6     \\
  \checkmark   &\checkmark              & \checkmark               &75.3      \\ \hline
\end{tabular}
}
\end{subtable}
\caption{Analysis of the multi-scale design of MPMC. 
}
\label{tab:ablscale}
\end{table*}

\noindent
\textbf{Analysis of Training Hyper-Parameters}
In this section, we analyze the training hyper-parameters used by MPMC,  which is integrated into UniMatch \cite{yang2023revisiting}.  The SSS pipeline is trained on the Cityscapes dataset's 1/16 and 1/8 data partitions.\\
\textbf{Analysis of $\alpha$ of the overall loss:} We analyze how MPMC performs with different values of $\alpha$, which is used to control the contribution of unsupervised segmentation loss in the overall loss (Equation $7$ of main paper). The results are provided in Table \ref{tab:tunable_paramter_alpha}. We observe $\alpha = 0.25$ achieves the best performance and is used in all our experiments
\begin{table}[hbt!]
%\small
  \centering
  \begin{tabular}{c|cc}
    \toprule
    %\multicolumn{2}{c}{Part}                   \\
    %\cmidrule(r){1-2}
   
$\alpha$ & 1/16 & 1/8 \\ 
 \midrule
   % \hline
0 & 66.3 & 72.8 \\
0.15 & 77.9 & 79.1 \\
0.20 & 78.3 & 79.2 \\
0.25 & \textbf{78.5} & \textbf{79.7}\\
0.30 & 77.9 & 79.0 \\
0.35 & 77.8 & 78.6\\
0.40 & 77.5 & 78.2\\
0.45 & 76.9 & 78.0\\

    \bottomrule
  \end{tabular}
  \caption{Analysis of ($\alpha$) of overall loss (1/16 and 1/8 partition protocols of Cityscapes Dataset) }
  \vspace{-4mm}
\label{tab:tunable_paramter_alpha}
\end{table}

\noindent
\textbf{Analysis of $\beta$ of the overall loss:} We analyze how MPMC performs with different values of $\beta$, which is used to control the contribution of unsupervised multi-label loss in the overall loss (Equation $7$ of main paper). The results are provided in Table \ref{tab:tunable_paramter_beta}. We observe that $\beta = 0.15$ achieves the best performance and is used in all our experiments\\
\begin{table}[hbt!]
%\small
  \centering
  \begin{tabular}{c|cc}
    \toprule
    %\multicolumn{2}{c}{Part}                   \\
    %\cmidrule(r){1-2}
   
$\beta$ & 1/16 & 1/8 \\ 
 \midrule
   % \hline
0 & 77.5 & 78.2 \\
0.05 & 77.7 & 78.8 \\
0.10 & 77.8 & 79.0 \\
0.15 & \textbf{78.5} & \textbf{79.7}\\
0.20 & 78.4 & 79.2 \\
0.25 & 78.2 & 78.9\\

    \bottomrule
  \end{tabular}
  \caption{Analysis of ($\beta$) of overall loss (1/16 and 1/8 partition protocols of Cityscapes Dataset) }
  \vspace{-4mm}
\label{tab:tunable_paramter_beta}
\end{table}

\noindent
\textbf{Analysis of per-class performance}
In Table \ref{tab:perclass}, we compare the per-class performance of our method incorporated into UniMatch \cite{yang2023revisiting} with respect to the baseline under $\frac{1}{16}$ cityscapes partition. We observe that our method improves the baseline across all classes. We find that MPMC using pseudo-label weighting, improves the performance of confusing classes like 
\textit{truck}, \textit{bus} and \textit{train} by $\mathbf{1.5}\%$, $\mathbf{1.4}\%$ and $\mathbf{2.5}\%$ respectively. These classes are often confused among each other because of sharing patch-based visual similarity and also having low training instances.

\begin{table*}[]
\begin{subtable}{1.0\linewidth}
\centering
\resizebox{1.0\linewidth}{!}{
\begin{tabular}{@{}ccccccccccccccccccccc@{}}
\toprule
\multicolumn{1}{c|}{Methods}       & \multicolumn{1}{c|}{\rotatebox{90}{mIoU}}  & \rotatebox{90}{road} & \rotatebox{90}{sidewalk} & \rotatebox{90}{building} & \rotatebox{90}{fence} & \rotatebox{90}{pole} & \rotatebox{90}{vegetation} & \rotatebox{90}{terrain} & \rotatebox{90}{sky} & \rotatebox{90}{person} & \rotatebox{90}{car} & \rotatebox{90}{wall} & \rotatebox{90}{traffic light} & \rotatebox{90}{traffic sign} & \rotatebox{90}{rider} & \rotatebox{90}{truck} & \rotatebox{90}{bus} & \rotatebox{90}{train} & \rotatebox{90}{motorcycle} & \rotatebox{90}{bicycle}\\ \midrule
\multicolumn{1}{c|}{UniMatch \cite{yang2023revisiting}}      & \multicolumn{1}{c|}{76.6}                &97.4      &79.9          &91.8          &57.9       &60.8      &92.3            &63.5         &94.6     &81.4        &94.0     &54.3      &69.9               &78.1              &61.5       &76.7       &85.1     &73.8   &66.2  &76.5    \\ \midrule
\multicolumn{1}{c|}{UniMatch+MPMC} & \multicolumn{1}{c|}{\textbf{78.5}}                 &\textbf{97.8}      &\textbf{80.1}          &\textbf{92.2}          &\textbf{58.2}       &\textbf{61.5}      &\textbf{92.8}            &\textbf{63.9}         &\textbf{95.1}     &\textbf{82.0}        &\textbf{94.5}     &\textbf{54.8}      &\textbf{71.0}               &\textbf{79.0}              &\textbf{63.4}       &\textbf{78.2}       &\textbf{86.5}     &\textbf{76.3}   &\textbf{68.2}  &\textbf{78.0}    \\ \bottomrule
\end{tabular}}
\end{subtable}
\caption{Per-class results on Cityscapes val set under $\frac{1}{16}$ data partition protocol}
\label{tab:perclass}
\end{table*}

\noindent
\textbf{Relationship between multi-label classifier confidence and noisy pseudo-labels}
In Equation $5$ of main paper, the per-class pseudo-pixel weight ($\lambda_s$) in a patch is estimated based on the corresponding per-class confidence of the proposed MPMC. This weight assignment is based on the hypothesis that a multi-label classifier is more confident (higher logits) for correct pseudo-labels than noisy pseudo-labels. The hypothesis is inspired from a recent study \cite{liu2020energy, wang2021can} that shows multi-label classifiers yield more confidence in prediction tasks for in-distribution data than out-of-distribution (OOD) data. 
Such a classifier property can be measured by label-wise energy scores . 
For a given input $x$ and the logits of a multi-label classifier $f(.)$, the label-wise energy score for each class $i$ is defined as:
\begin{equation}\label{eq13}
     E_{i}(x) = \log(1 + e^{f_{i}(x)})
\end{equation}
Consequently, a high label-wise energy score for class $i$ corresponds to high class logit (confidence) and vice-versa. Our key observation is that the behavior of a multi-label classifier, in this case, MPMC, towards in-distribution and Out-of-Distribution (OOD) data, also applies to  correct and noisy pseudo-labels. 

\begin{figure*}
\centering
\includegraphics[width=0.95\linewidth,height=33\baselineskip]{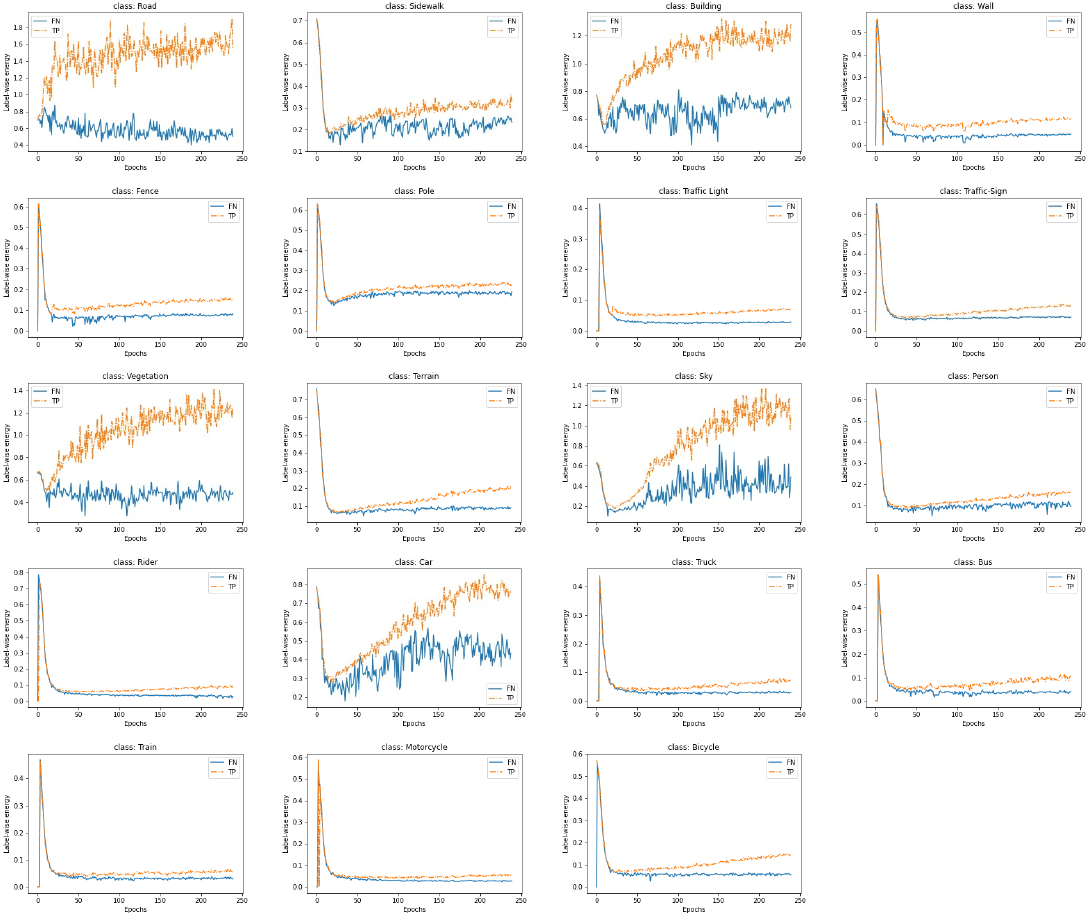}
% \vspace{0.5cm}
\caption{Label-wise energy scores for true positive and false negative for all classes in the Cityscapes validation set.  
}
  \label{fig:graphsupp}
\end{figure*}

To corroborate the above hypothesis, we trained a segmentation network with ResNet-101 as the backbone and DeepLabV3+ as the decoder, using MPMC exclusively on labeled images from a $\frac{1}{16}$ partition of the Cityscapes dataset. We then analyzed the label-wise energy scores for true positives (correct pseudo-labels) and false negatives (noisy pseudo-labels) in the validation set. The results, as shown in \cref{fig:graphsupp}, indicate that MPMC consistently registers lower label-wise energy scores for noisy pseudo-labels (false negatives) across all classes compared to correct pseudo-labels (true positives) following the initial training epochs.

\noindent
\textbf{Ablation study on the plug-in position of classifier}
In Table \ref{tab:encposition}, we analyse the influence of the classifier in different positions of the segmentation network on Pascal VOC dataset. We observe that the highest performance boost occurs when the classifier is inserted at the first layer, which preserves access to low-level information and edges. 
\begin{table}[hbt!] 
\begin{subtable}{\columnwidth}
\centering
\resizebox{0.85\columnwidth}{!}{%
\begin{tabular}{l|c|c|c|c|c}
\hline
     & \begin{tabular}[c]{@{}c@{}}Enc. Layer1\end{tabular} & \begin{tabular}[c]{@{}c@{}}Enc. Layer2\end{tabular} & \begin{tabular}[c]{@{}c@{}}Enc. Layer3\end{tabular} & \begin{tabular}[c]{@{}c@{}}Enc. Layer4\end{tabular} & Decoder End \\ \hline
mIoU & \textbf{77.3}                                               & 77.0                                                        & 76.9                                                        & 76.4                                                        & 75.8        \\ \hline
\end{tabular}}
\end{subtable}
\caption{Effect of plug-in position of MPMC in the encoder 
}

%\caption{%Analysis of influence of the classifier in different positions of the segmentation network for Pascal VOC dataset
%}
\label{tab:encposition}
\end{table}

\noindent
\textbf{Performance gains with Vision Transformers (ViT)} 
In our approach, we employ AugSeg \cite{zhao2023augmentation}, a semi-supervised segmentation method, and modify its segmentation network by integrating Segmenter \cite{strudel2021segmenter}, which utilizes a ViT as the encoder. As observed in Table \ref{tab:vit} below for Cityscapes dataset, MPMC enhances the performance of the ViT-based model across all data partitions. This improvement indicates that our MPMC introduces complementary information, distinct from the patch processing in ViTs. Note that in low data regimes, convolution-based methods surpass transformer-based approaches in performance. 

\begin{figure*}
\centering
\includegraphics[width=0.65\linewidth,]{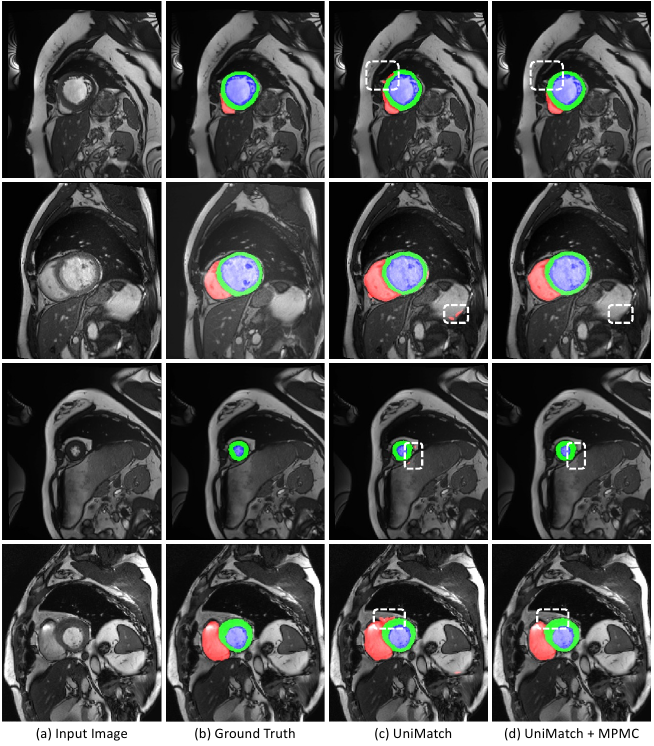}
% \vspace{0.5cm}
\caption{\textbf{Qualitative Results on ACDC dataset:} (a) input image, (b) ground truth, (c) segmentations generated by UniMatch \cite{yang2023revisiting} compared to (d) which are segmentations generated when our method (MPMC) is integrated to UniMatch. The white boxes show the areas where our method improves the baseline \cite{yang2023revisiting}.  
}\label{fig:unimatchsupacdcpic}
\end{figure*}

\begin{figure*}
\centering
\includegraphics[width=0.65\linewidth,]{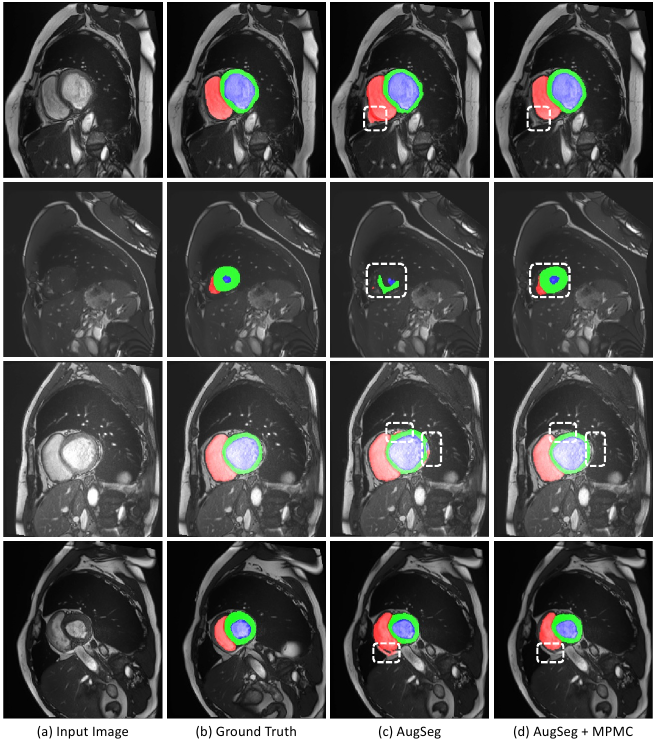}
% \vspace{0.5cm}
\caption{\textbf{Qualitative Results on ACDC dataset:} (a) input image, (b) ground truth, (c) segmentations generated by AugSeg \cite{zhao2023augmentation} compared to (d) which are segmentations generated when our method (MPMC) is integrated to AugSeg. The white boxes show the areas where our method improves the baseline \cite{zhao2023augmentation}.  
}\label{fig:augsegsupacdcpic}
\end{figure*}

\begin{figure*}
\centering
\includegraphics[width=1\linewidth,]{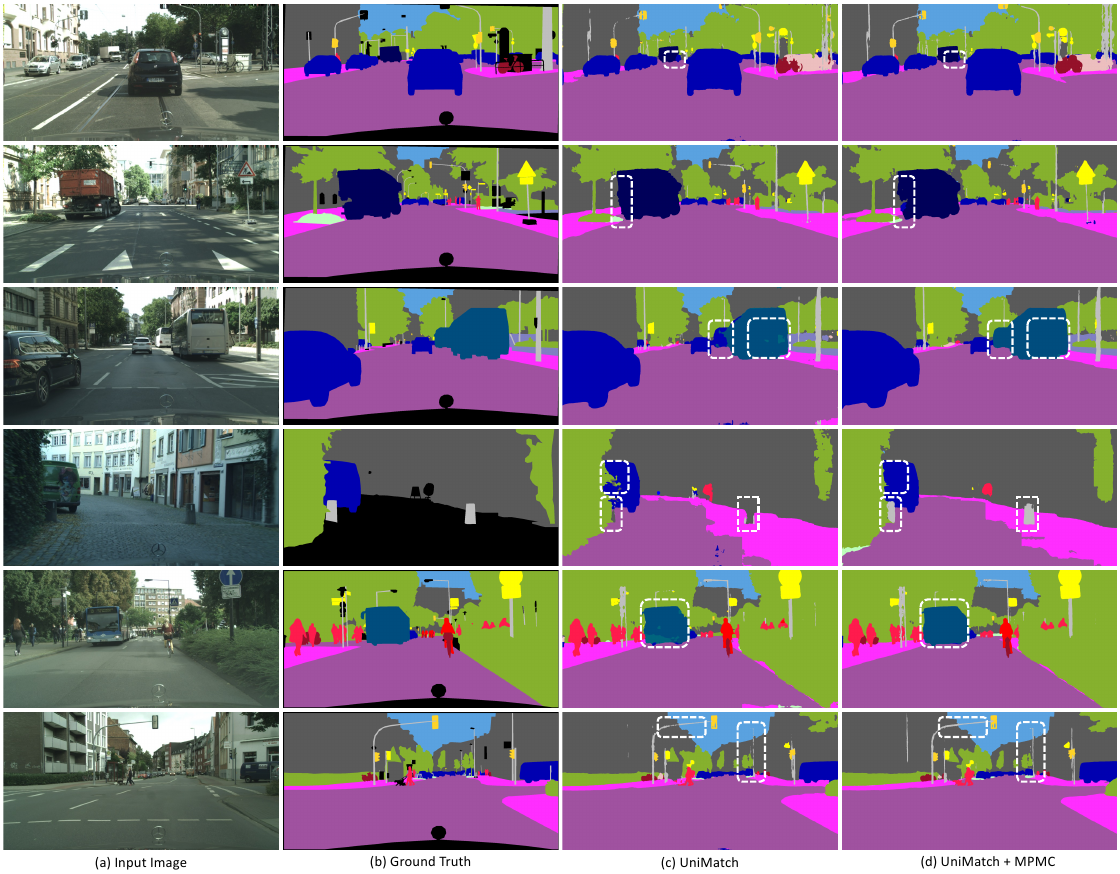}
% \vspace{0.5cm}
\caption{\textbf{Qualitative Results on Cityscapes dataset:} (a) input image, (b) ground truth, (c) segmentations generated by UniMatch \cite{yang2023revisiting} compared to (d) which are segmentations generated when our method (MPMC) is integrated to UniMatch. The white boxes show the areas where our method improves the baseline \cite{yang2023revisiting}.  
}
  \label{fig:unimatchsupcitypic}
\end{figure*}

\begin{figure*}
\centering
\includegraphics[width=1\linewidth,]{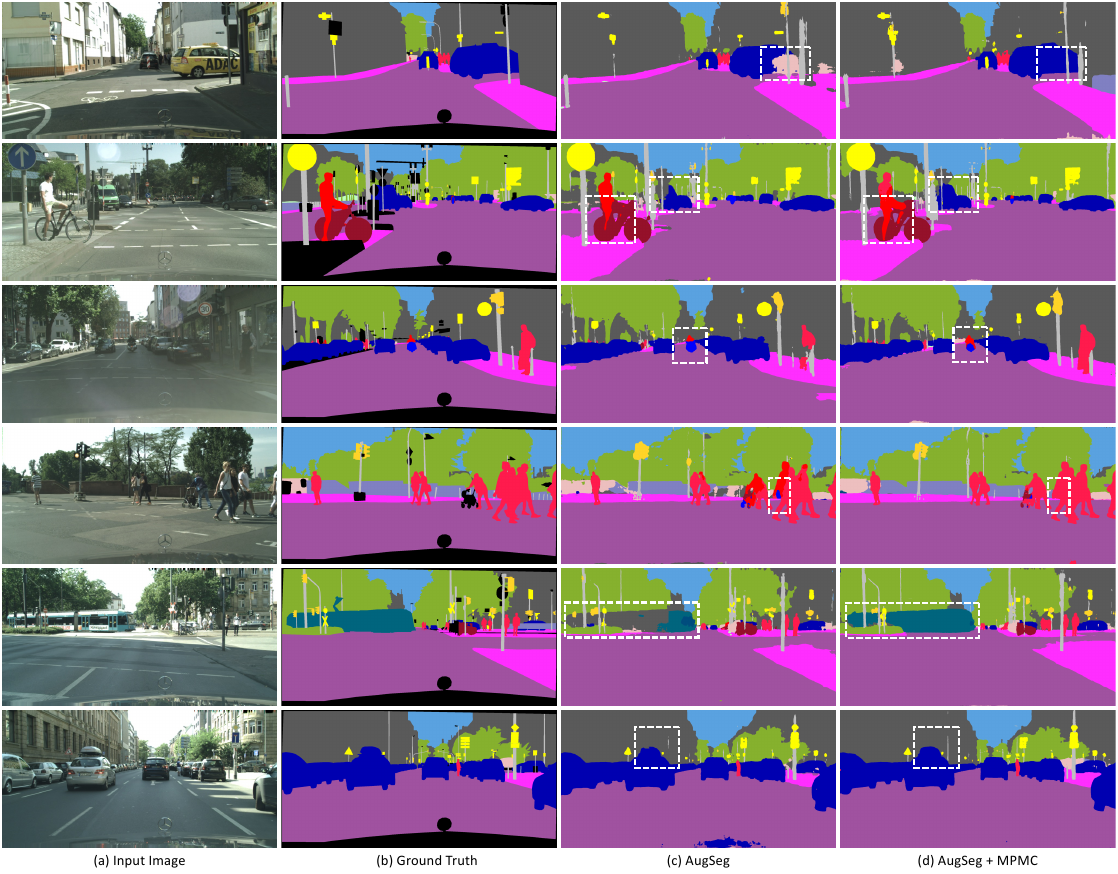}
% \vspace{0.5cm}
\caption{\textbf{Qualitative Results on Cityscapes dataset:} (a) input image, (b) ground truth, (c) segmentations generated by AugSeg \cite{zhao2023augmentation} compared to (d) which are segmentations generated when our method (MPMC) is integrated to AugSeg. The white boxes show the areas where our method improves the baseline \cite{zhao2023augmentation}.  
}
  \label{fig:augsegsupcitypic}
\end{figure*}

\begin{figure*}
\centering
\includegraphics[width=1\linewidth,]{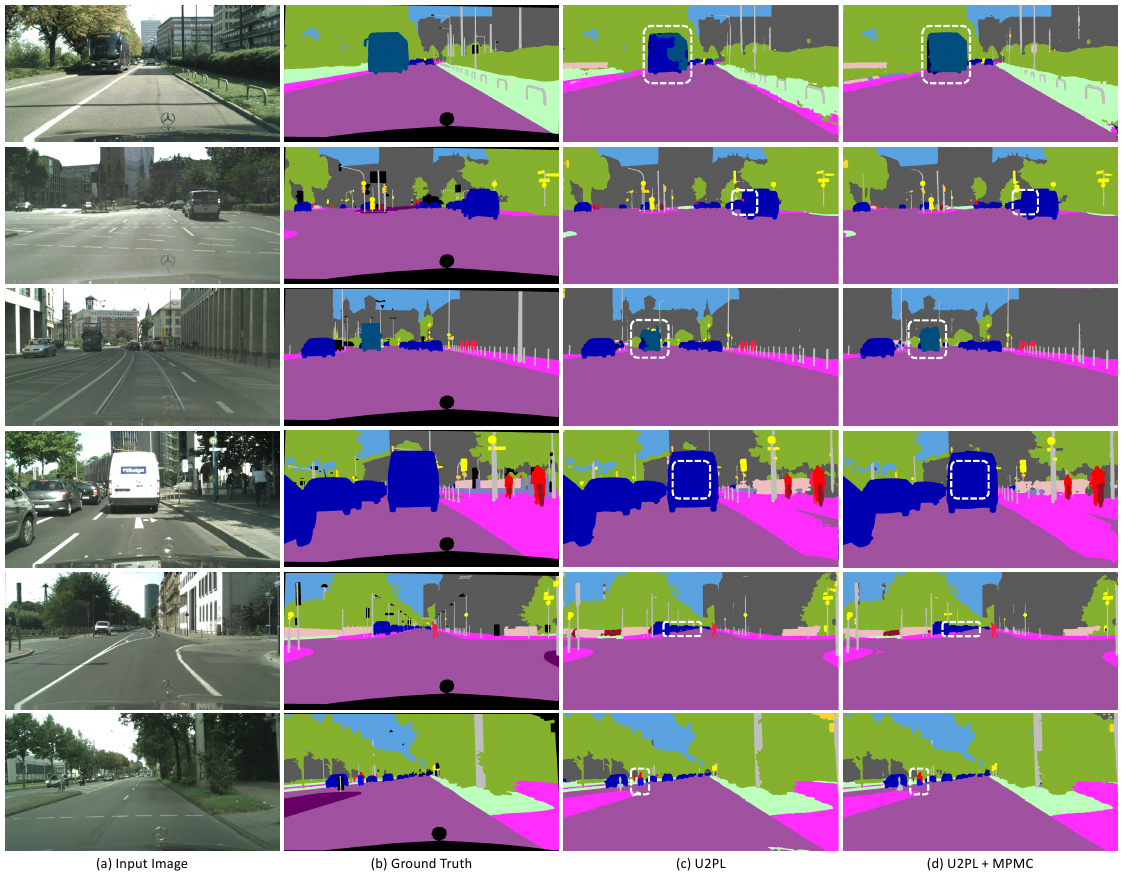}
% \vspace{0.5cm}
\caption{\textbf{Qualitative Results on Cityscapes dataset:} (a) input image, (b) ground truth, (c) segmentations generated by U2PL \cite{wang2022semi} compared to (d) which are segmentations generated when our method (MPMC) is integrated to U2PL. The white boxes show the areas where our method improves the baseline \cite{wang2022semi}.  
}
  \label{fig:u2plsupcitypic}
\end{figure*}

\begin{figure*}
\centering
\includegraphics[width=1\linewidth,]{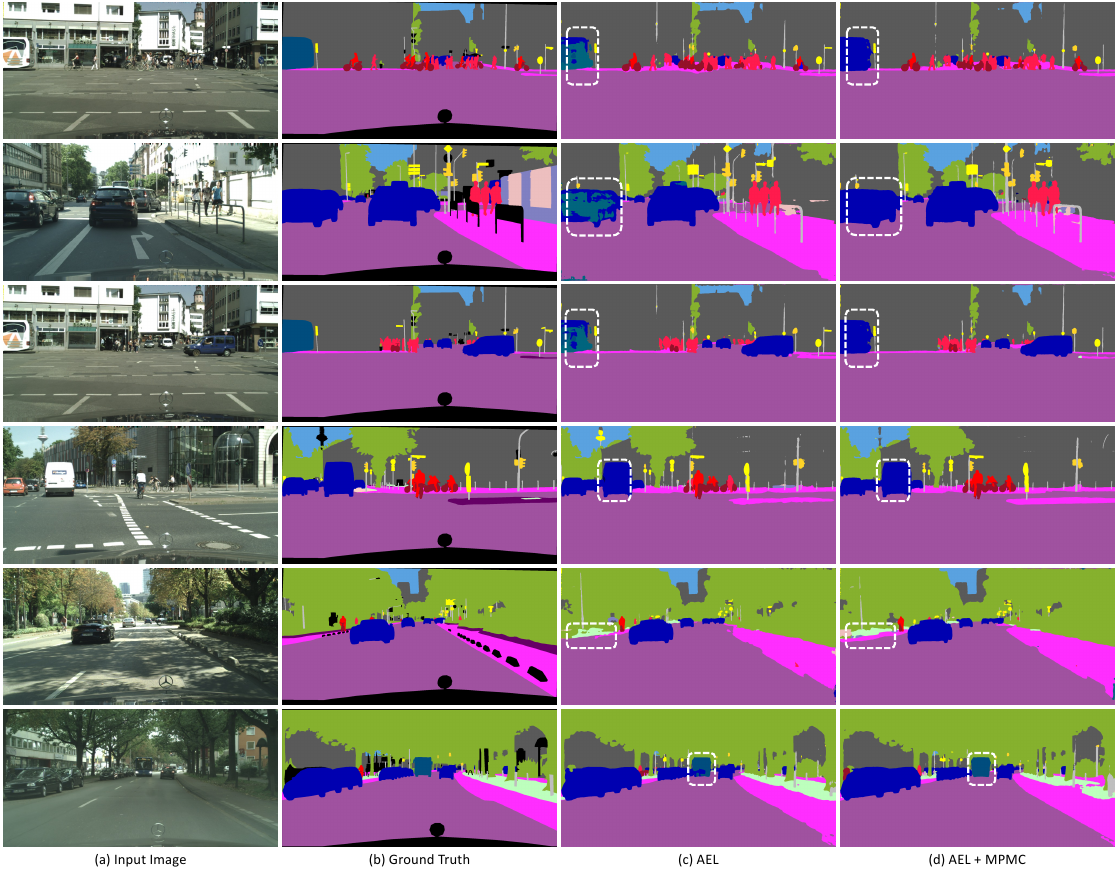}
% \vspace{0.5cm}
\caption{\textbf{Qualitative Results on Cityscapes dataset:} (a) input image, (b) ground truth, (c) segmentations generated by AEL \cite{hu2021semi} compared to (d) which are segmentations generated when our method (MPMC) is integrated to AEL. The white boxes show the areas where our method improves the baseline \cite{hu2021semi}.  
}
  \label{fig:aelsupcitypic}
\end{figure*}

\begin{figure*}
\centering
\includegraphics[width=1\linewidth,]{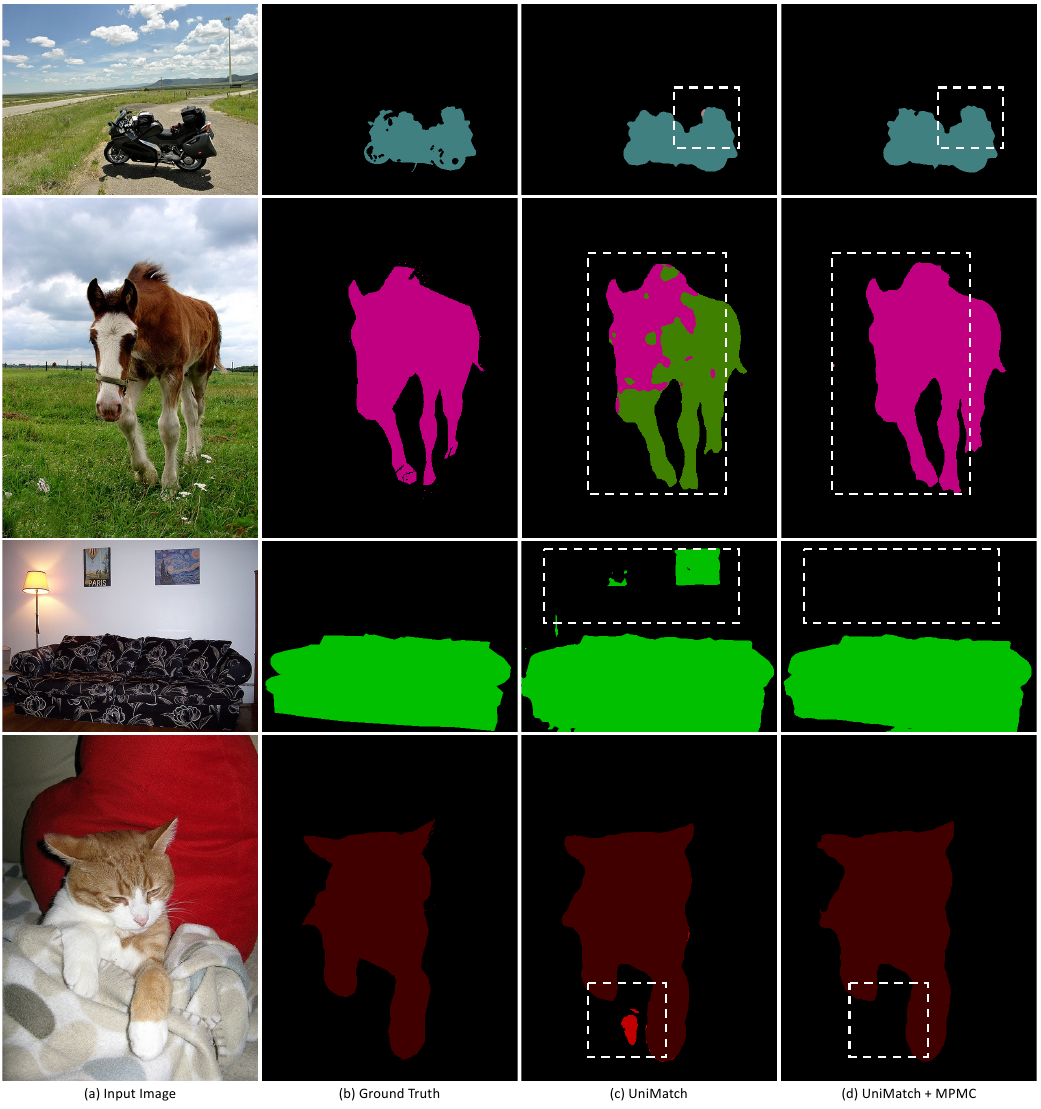}
% \vspace{0.5cm}
\caption{\textbf{Qualitative Results on Pascal VOC dataset:} (a) input image, (b) ground truth, (c) segmentations generated by UniMatch \cite{yang2023revisiting} compared to (d) which are segmentations generated when our method (MPMC) is integrated to UniMatch. The white boxes show the areas where our method improves the baseline \cite{yang2023revisiting}.  
}
  \label{fig:unimatchsuppascalpic}
\end{figure*}

\begin{figure*}
\centering
\includegraphics[width=1\linewidth,]{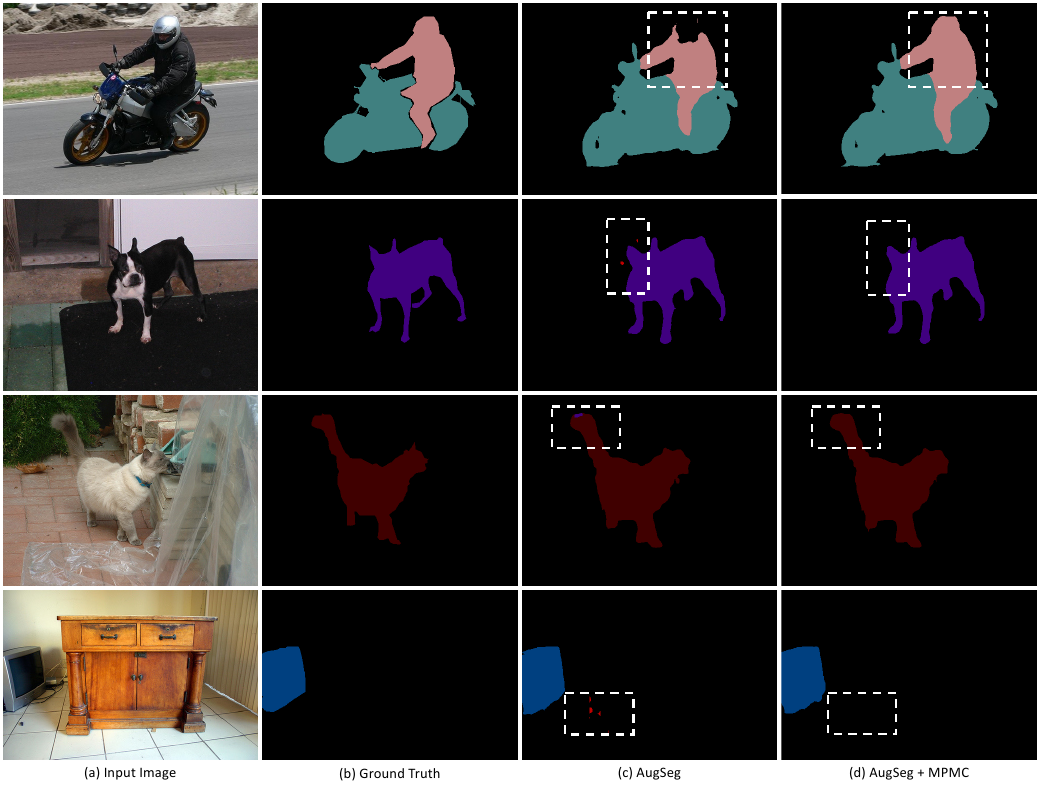}
% \vspace{0.5cm}
\caption{\textbf{Qualitative Results on Pascal VOC dataset:} (a) input image, (b) ground truth, (c) segmentations generated by AugSeg \cite{zhao2023augmentation} compared to (d) which are segmentations generated when our method (MPMC) is integrated to AugSeg. The white boxes show the areas where our method improves the baseline \cite{zhao2023augmentation}.  
}
  \label{fig:augsegsuppascalpic}
\end{figure*}

\begin{table}[hbt!]
\begin{subtable}{\columnwidth}
\centering
\resizebox{0.7\columnwidth}{!}{%
\begin{tabular}{c|c|c|c|c|c}
\hline
Method       & Backbone & 1/16          & 1/8           & 1/4           & 1/2  \\ \hline
AugSeg (Seg-L)  & ViT-L     & 56.6          & 60.2          & 65.8         & 71.3 \\
AugSeg (Seg-L) + MPMC & ViT-L  & \textbf{58.0} & \textbf{61.2} & \textbf{66.7} &  \textbf{72.3}    \\ \hline

\end{tabular}}
\end{subtable}
\caption{Effect MPMC in ViT-based Architecture}
\label{tab:vit}
\end{table}
%%%%%%%%%%%%%%%%%%%%%%%
% \textbf{Confidence of Multi-label classifiers vs semantic segmentation head}
\noindent
\textbf{Qualitative Results}
\begin{itemize}

\item \textbf{ACDC}: In \cref{fig:unimatchsupacdcpic} and \cref{fig:augsegsupacdcpic},  we compare our method (MPMC) integrated into UniMatch \cite{yang2023revisiting} and AugSeg \cite{zhao2023augmentation} respectively, with the corresponding baselines (UniMatch and AugSeg). Through visualizing the segmentation results, we observe that our method reduces the number of pixels being misclassified for both UniMatch \cite{yang2023revisiting} and AugSeg \cite{zhao2023augmentation}.

Note, all methods have been trained on ``3 cases'' data partition which has the least labeled images ($<10\%$ of labeled images).

\item \textbf{Cityscapes}: In \cref{fig:unimatchsupcitypic}, \cref{fig:augsegsupcitypic}, \cref{fig:u2plsupcitypic} and \cref{fig:aelsupcitypic} we compare our method (MPMC) integrated into UniMatch \cite{yang2023revisiting}, AugSeg \cite{zhao2023augmentation}, U2PL \cite{wang2022semi} and AEL \cite{hu2021semi} respectively, with the corresponding baselines (UniMatch, AugSeg, U2PL and AEL). Through visualizing the segmentation results, we observe that our method reduces the number of pixels being misclassified for all four baselines.

Note, all methods have been trained on $\frac{1}{16}$ data partition of Cityscapes dataset and all visualizations are on  Cityscapes validation set. 

\item \textbf{Pascal VOC}: In \cref{fig:unimatchsuppascalpic} and \cref{fig:augsegsuppascalpic},  we compare our method (MPMC) integrated into UniMatch \cite{yang2023revisiting} and AugSeg \cite{zhao2023augmentation} respectively, with the corresponding baselines (UniMatch and AugSeg). Through visualizing the segmentation results, we observe that our method reduces the number of pixels being misclassified for both UniMatch \cite{yang2023revisiting} and AugSeg \cite{zhao2023augmentation}.

Note, all methods have been trained on $\frac{1}{16}$ data partition of Pascal VOC dataset (\textit{Classic}) and all visualizations are on Pascal VOC validation set. 

\end{itemize}

%%%%%%%%%%%%%%%%%%%%%%%%%%%%%%%%%%%%%%%%%

%%%%%%%%%%%%%%%%%%%%%%%%%%%%%%%%%%%
%%%%%%%%%%%%%%%%%%%%%%%%%%%%%%%%%%%%

%%%%%%%%%%%%%%%%%%%%%

\end{document}